\documentclass[acmsmall,screen]{acmart}

\AtBeginDocument{%
	\providecommand\BibTeX{{%
			\normalfont B\kern-0.5em{\scshape i\kern-0.25em b}\kern-0.8em\TeX}}}

\usepackage{makecell}
\usepackage{multirow}
\usepackage{url}
\begin{document}
	
	\title{A Systematic Survey of Regularization and Normalization in GANs}
	
	\author{Ziqiang Li}
	\email{iceli@mail.ustc.edu.cn}
	\author{Muhammad Usman}
	\email{muhammadusman@mail.ustc.edu.cn}
	\author{Rentuo Tao}
	\email{trtmelon@mail.ustc.edu.cn}
	\author{Pengfei Xia}
	\email{xpengfei@mail.ustc.edu.cn}
	\affiliation{%
		\institution{ University of Science and Technology of China}
		\country{China}
	}
	\author{Chaoyue Wang}
	\email{chaoyue.wang@outlook.com}
	\affiliation{%
		\institution{JD Explore Academy}
		\country{China}
	}
	
	\author{Huanhuan Chen}
	\authornote{corresponding author}
	\email{hchen@ustc.edu.cn}
	\author{Bin Li}
	\email{binli@ustc.edu.cn}
	\authornotemark[1]
	\affiliation{%
		\institution{ University of Science and Technology of China}
		\country{China}
	}

	\renewcommand{\shortauthors}{Submitted to ACM Computing Surveys.}
	
	\begin{abstract}
		Generative Adversarial Networks (GANs) have been widely applied in different scenarios thanks to the development of deep neural networks. The original GAN was proposed based on the non-parametric assumption of the infinite capacity of networks. However, it is still unknown whether GANs can fit the target distribution without any prior information. Due to the overconfident assumption, many issues remain unaddressed in GANs' training, such as non-convergence, mode collapses, gradient vanishing. Regularization and normalization are common methods of introducing prior information to stabilize training and improve discrimination. Although a handful number of regularization and normalization methods have been proposed for GANs, to the best of our knowledge, there exists no comprehensive survey which primarily focuses on objectives and development of these methods, apart from some in-comprehensive and limited scope studies. In this work, we conduct a comprehensive survey on the regularization and normalization techniques from different perspectives of GANs training. First, we systematically describe different perspectives of GANs training and thus obtain the different objectives of regularization and normalization. Based on these objectives, we propose a new taxonomy. Furthermore, we compare the performance of the mainstream methods on different datasets and investigate the applications of regularization and normalization techniques that have been frequently employed in state-of-the-art GANs. Finally, we highlight potential future directions of research in this domain. Code and studies related to the regularization and normalization of GANs in this work is summarized on https://github.com/iceli1007/GANs-Regularization-Review. 
		
	\end{abstract}

	\ccsdesc[500]{Computing methodologies~Computer vision}
	\ccsdesc[500]{Computing methodologies~Regularization}
	\ccsdesc[500]{Computing methodologies~Neural networks}
	\keywords{Generative Adversarial Networks, Lipschitz Neural networks, Training Dynamics}
	\setcopyright{acmcopyright}
	\acmJournal{CSUR}
	\acmYear{2022} \acmVolume{1} \acmNumber{1} \acmArticle{1} \acmMonth{1} \acmPrice{15.00}\acmDOI{10.1145/3569928}
	
	\maketitle
	
	\section{Introduction}
	\setlength{\abovecaptionskip}{0.cm}
	%
	%
	%
	%
	Generative adversarial networks (GANs) \cite{goodfellow2014generative} have been widely used in computer vision, such as image inpainting \cite{yu2018generative,demir2018patch,javed2019image,wang2018perceptual,wang2017tag}, style transfer \cite{gonzalez2018image,royer2020xgan,lee2020drit++,choi2020stargan}, text-to-image translations \cite{zhang2017stackgan,xu2018attngan,qiao2019mirrorgan}, and attribute editing \cite{shen2020interpreting,choi2018stargan,chen2020puppeteergan,tao2019resattr,9399843,9412045}. GANs training is a two-player zero-sum game between a generator and a discriminator, which can be understood from different perspectives: (i) "Real \& Fake" \cite{goodfellow2014generative,mao2017least}, (ii) "Fitting distribution" \cite{nowozin2016f,arjovsky2017wasserstein}, and (iii) "Training dynamics" \cite{mescheder2017numerics, heusel2017gans}. GANs training suffers from several issues, for instance: non-convergence \cite{kodali2017convergence,nie2019towards}, mode collapses \cite{srivastava2017veegan}, gradient vanishing \cite{Arjovsky2017Towards}, overfitting \cite{yazici2020empirical}, discriminator forgetting \cite{chen2019self} and deficiency \cite{chen2020ssd}, and hyperparameters sensitivity \cite{kurach2018gan}. Many solutions to mitigate these issues have been proposed, focusing on designing new architectures \cite{karras2017progressive,brock2018large}, loss functions \cite{arjovsky2017wasserstein,gulrajani2017improved,chen2019self,zhao2020improved,wang2019evolutionary}, optimization methods \cite{heusel2017gans,brock2018large}, regularization \cite{gulrajani2017improved,mescheder2017numerics}, and normalization \cite{miyato2018spectral}. Among them, regularization and normalization techniques are compatible with loss functions, model structures, and tasks, which has attracted the attention of scholars. 
	
	Regularization and normalization are widely applied in neural networks training to introduce prior knowledge. For supervised tasks, regularization in literature has been proposed to introduce some advantages like overfitting prevention \cite{park2008bayesian,hoerl1970ridge}, semi-supervised assumptions \cite{soares2012semisupervised}, manifold assumptions \cite{hong2019variant,zhao2020semi}, feature selection \cite{jiang2019probabilistic}, and low rank representation \cite{hu2020low}. On the other hand, normalization \cite{ioffe2015batch,ba2016layer} is advantageous for the Stochastic Gradient Descent (SGD) \cite{bottou2010large}, accelerating convergence and improving accuracy. Unlike the icing on the cake of supervisory tasks, regularization and normalization are utilized inevitably in weak-supervised and unsupervised tasks. GANs' training is a two-player zero-sum game having a solution to Nash equilibrium. The proposal of standard GAN is based upon the non-parametric assumption of the infinite capacity of networks, an unsupervised learning task. Likewise, a good number of research studies targeting GANs training from different perspectives argue that unconstrained training causes unstable training (generator \cite{wu2020improving} and discriminator \cite{gulrajani2017improved}) and significant bias between real images and fake images (attributes domain \cite{zhao2018bias} and frequency domain \cite{chen2020ssd,li2021high}). Therefore, a large amount of prior should be introduced into GANs training through regularization and normalization
	
	\begin{figure}
		\centering
		\includegraphics[scale=.52]{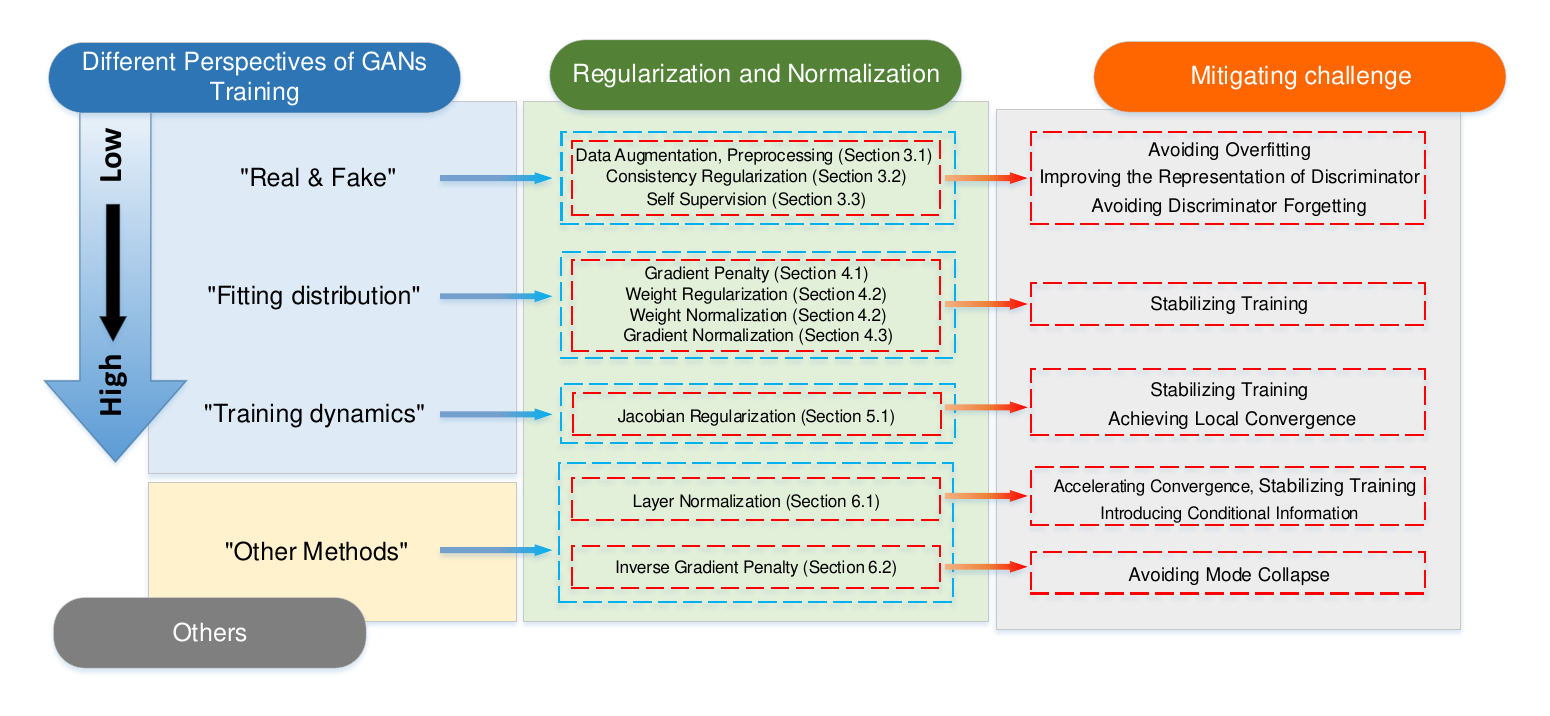}
		\caption{The overview of different perspectives of GANs training and the summary of the regularization and normalization for GANs.}
		\label{fig:overview}
	\end{figure}
	
	Regularization and normalization are effectively used to stabilize training and improve the performance of GANs in existing literature \cite{gulrajani2017improved,petzka2017regularization,wei2018improving}. Due to diverse nature of the topic, there is a need for systematic literature survey. There exist some literature studies \cite{kurach2018large,kurach2018gan,lee2020regularization}, however, these studies either lack comprehensive coverage of the topic, miss detailed background information and theoretical analysis, or do not correlate different methods. In this paper, based on the different perspectives of GANs training, we propose a new taxonomy, denoted as \textbf{"Real \& Fake"}, \textbf{"Fitting distribution"}, \textbf{"Training dynamics"}, and \textbf{"Other methods"}, for a better understanding of regularization and normalization during the GANs training as depicted in Figure \ref{fig:overview}. {\textbf{"Real \& Fake"} is the low-level (intuitive) perspective of GANs, in which GANs is considered as "counterfeiters-police" competition. At this level, D estimates the real probability of both real and fake samples, which is similar to the bi-classification task. Therefore, prior information and additional supervision tasks in classification task are also urgent during the training process of the discriminator. Based on these, some regularization methods, such as {\textit{Data Augmentation and Preprocessing}}}\footnote{Data augmentation and preprocess introduce additional data and prior, which is similar to regularization. More importantly, both consistency regularization and self-supervision need different data transformation operations. Hence, this paper also discusses some works on this.}, {\textit{Consistency Regularization}, and {\textit{Self-Supervision}} are proposed to mitigate overfitting, improve the representation of discriminator, and avoid discriminator forgetting by introducing additional supervised information and data; \textbf{"Fitting distribution"} is the middle-level perspective of GANs. At this level, generator is considered as a distribution mapping function, and the discriminator is a distribution divergence. Among various distances, Wasserstein distance is a popular form, and Lipschitz continuity is a necessary condition for achieving Wasserstein distance. Based on these, \textit{Gradient Penalty}, \textit{Weight Normalization}, \textit{Weight Regularization}, and  \textit{Gradient} \textit{Normalization} are used to fulfill Lipschitz continuity and ensure training stability of discriminator; \textbf{"Training Dynamics"} is the high-level (essential) perspective of GANs. At this level, GANs training is a two-player zero-sum game with a solution to Nash Equilibrium. To achieve theoretical local convergence, \textit{Jacobian Regularization} needs to be used; Finally, \textbf{"Other methods"} containing \textit{Layer} \textit{ Normalization} and \textit{Inverse Gradient Penalty} are used for conditional generation and easing mode collapse, respectively}

	In summary, we make the following contributions in this survey:
	\begin{itemize}
		\item \textit{Comprehensive analysis of GANs training.} In this study, we analyze the GANs training from three perspectives including "Real \& Fake", "Fitting distribution", and "Training dynamics". To the best of our knowledge, this survey is the first in this domain with comprehensive analysis.
		
		\item \textit{New taxonomy.} Based on the analysis of GANs training from different perspectives, we propose a novel taxonomy and contextualize the regularization and normalization of GANs comprehensively.
		
		\item \textit{ Comparison and analysis.} Following the taxonomy, we also provide quantitative and qualitative analysis and comparison for each type of regularization and normalization techniques, which has helped the researchers and practitioners navigate this space.
	\end{itemize}
	
	{\textbf{The Scope of This Survey.} This survey aims to systematically analyze the prevalent problems in the GANs training, such as non-convergence, mode collapses, gradient vanishing, and discriminator overfitting. Accordingly, different regularization and normalization technologies have been summarized. Of course, not all regularization and normalization technologies for GANs are covered in this survey. In some case, some regularization and normalization technologies are highly dependent on the task and data in the hand, we recommend looking for domain-specific regularization and normalization techniques from the following reviews: data-efficient generation \cite{li2022comprehensive}, medical image generation\cite{singh2021medical}\cite{yi2019generative}, image super-resolution \cite{tian2022generative}, biomedical informatics generation \cite{lan2020generative}, Spatio- temporal data generation \cite{gao2022generative}, text generation \cite{de2021survey}. Our survey is concerned with general technologies in GANs, which are not dependent on model structures, data, and task. We hope our study can provide general and universal insights of GANs for the community.}
	
	The rest of this paper is organized as follows: Section 2 introduces the background and different training perspectives of GANs. Section 3, 4, 5, and 6
	describe regularization and normalization methods in different groups, respectively. Furthermore, we investigate the applications of regularization and normalization techniques that have been frequently employed in SOTA GANs in Section 7 and discuss the current problems and prospects for future work in Section 8.

	\section{Background and Three Perspectives of GANs Training}
	\subsection{Regularization and Normalization}
	{Regularization and normalization are common and important techniques to introduce prior knowledge in neural networks.} Regularization is a technique to control the complexity of learning models. Weight decay \cite{krogh1992simple} is a typical method to minimize the square of weights together with the training loss in the training of neural networks \cite{chen2008diversity,kukavcka2017regularization}, which can be used to improve generalization. 
	In Bayesian learning methods, such as relevance vector machine \cite{tipping2001sparse}, probabilistic classification vector machines \cite{chen2009probabilistic,chen2013efficient}, and others  \cite{lyu2019multiclass}, regularization is termed as prior distribution. Specifically, L2 regularization \cite{hoerl1970ridge} is equivalent to introducing Gaussian prior to the parameters, and L1 regularization \cite{park2008bayesian} is equivalent to introducing Laplace prior to the parameters. The theoretical connection between regularization and prior information has been investigated in neural network ensembles research \cite{chen2009regularized,chen2010multiobjective}. Regularization does not only control overfitting but also provide other characteristics like semi-supervised assumptions \cite{soares2012semisupervised}, manifold assumptions \cite{hong2019variant,zhao2020semi}, feature selection \cite{jiang2019probabilistic}, low rank representation \cite{hu2020low,wen2018low}, and consistency assumptions \cite{zhang2019consistency, zhou2019discriminative}. {Normalization \cite{ioffe2015batch,ba2016layer} is the mapping of data to a specified range, which is advantageous for the Stochastic Gradient Descent (SGD) \cite{bottou2010large}, accelerating convergence and improving accuracy.}
	
	\subsection{GANs}
	GANs are two-player zero-sum games, where generator (G) and discriminator ($D$) try to optimize opposing loss functions to find the global Nash equilibrium. In general, GANs can be formulated as follows:
	\begin{equation}
		\mathop{\min}\limits_{\phi}\mathop{\max}\limits_{\theta} f(\phi,\theta)
		=\mathop{\min}\limits_{\phi}\mathop{\max}\limits_{\theta}\mathbb{E}_{x\sim p_{r}}[g_1(D_\theta(x))]
		+\mathbb{E}_{z\sim p_z}[g_2(D_\theta(G_\phi(z)))],
		\label{EQ:eqn1}
	\end{equation}
	where $\phi$ and $\theta$ are parameters of the generator $G$ and the discriminator $D$, respectively. $p_r$ and $p_z$ represent the real distribution and the latent distribution, respectively. {$g_1$and $g_2$ are different functions corresponding to different GANs. Specifically, vanilla GAN \cite{goodfellow2014generative} can be described as $g_1(t)=g_2(-t)=-\log(1+e^{-t})$; $f$-GAN \cite{nowozin2016f} can be written as $g_1(t)=-e^{-t}$, $g_2(t)=1-t$; Morever, Geometric GAN \cite{lim2017geometric} and WGAN \cite{arjovsky2017wasserstein} are described as $g_1(t)=g_2(-t)=-\mathop{\max}(0,1-t)$ and $g_1(t)$\\$=g_2(-t)=t$, respectively.} 
	
	{Different from supervised learning, GANs training is an unsupervised learning, which leads to the urgency of regularization and normalization in the training of GANs.} In the following parts, we elaborate the training of GANs from three perspectives: low level: the perspective of "Real \& Fake", middle level: the perspective of "Fitting distribution", and high level: the perspective of "Training dynamics". According to different perspectives of GANs, various regularization and normalization have been proposed in the GANs training.

	\subsection{Low Level: The Perspective of "Real \& Fake"}
	In low level, GANs is considered as "counterfeiters-police" competition, where the generator (G) can be thought of counterfeiters, trying to produce fake currency and use it undetected, while the discriminator (D) is analogous to the police, trying to detect the counterfeit currency. This competition drives both teams to upgrade their methods until the fake currency is indistinguishable from the real ones. Generally, D estimates the real probability of both real and fake samples, which is very similar to the bi-classification task, while G generates fake samples similar to real ones. Hence, the loss function in Eq (\ref{EQ:eqn1}) is formulated as:
	
	\begin{equation}
		\mathop{\min}\limits_{\phi}\mathop{\max}\limits_{\theta} f(\phi,\theta)
		=\mathop{\min}\limits_{\phi}\mathop{\max}\limits_{\theta}\mathbb{E}_{x\sim p_{r}}[\log(D_\theta(x))]
		+\mathbb{E}_{z\sim p_z}[\log(1-D_\theta(G_\phi(z)))],
		\label{EQ:eqnganlevel1}
	\end{equation}
	where $f(\phi,\theta)$ is a binary cross-entropy function, commonly used in binary classification problems. Eq (\ref{EQ:eqnganlevel1}) is proposed in original GAN \cite{goodfellow2014generative} and can be optimized by alternate training. The training of discriminator is:
	\begin{equation}
		\mathop{\max}\limits_{\theta} \mathbb{E}_{x\sim p_{r}}[\log(D_\theta(x))]
		+\mathbb{E}_{z\sim p_z}[\log(1-D_\theta(G_\phi(z)))],
		\label{EQ:eqndiscriminator}
	\end{equation}
	which is the same as a bi-classification task between real images and generated images. However, the naive binary cross-entropy function suffers from many problems, such as gradients vanishing. {Gradients vanishing is present when the difference between real and generated images as measured by discriminator is large, which leads generators cannot get optimised directions.} Accordingly, many techniques from classification like loss functions and regularization methods have been used to improve the training of discriminator. For instance, to overcome the gradients vanishing problem, Mao et al. \cite{mao2017least} propose the LSGANs which adopts the least
	squares loss function for the discriminator. The least squares loss function moves the fake samples toward the decision boundary even though they are correctly classified. Based on this property, LSGANs is able to generate samples that are closer to real ones. The loss functions of LSGANs can be defined as follows:
	\begin{equation}
		\begin{aligned}
			&\mathop{\min}\limits_{\theta} \mathbb{E}_{x \sim p_{r}}[(D_{\theta}({x})-b)^{2}] 
			+ \mathbb{E}_{{z} \sim p_{{z}}}[(D_{\theta}(G_{\phi}({z}))-a)^{2}], \\
			&\mathop{\min}\limits_{\phi}  \mathbb{E}_{z \sim p_{z}}[(D_{\theta}(G_{\phi}({z}))-c)^{2}],
		\end{aligned}
	\end{equation}
	where $b$ and $a$ are objectives that $D$ uses for the training of real and fake samples respectively, $c$ denotes the value that $G$ wants $D$ to believe for fake sample. Gradients vanishing problem of the LSGANs only appears with $D_{\theta}(G_{\phi}({z}))=c$, which is hard. Furthermore, Lin et al. \cite{lim2017geometric} use SVM separating hyperplane that maximizes the margin to propose geometric GAN. Authors use the Hinge loss to train the models, which can be formulated as:
	\begin{equation}
		\begin{aligned}
			&\mathop{\min}\limits_{\theta} \mathbb{E}_{x \sim p_{r}}[(1-D_{\theta}({x})]_{+}
			+ \mathbb{E}_{{z} \sim p_{{z}}}[1+D_{\theta}(G_{\phi}({z}))]_{+}, \\
			&\mathop{\min}\limits_{\phi}  -\mathbb{E}_{z \sim p_{z}}D_{\theta}(G_{\phi}({z})),
		\end{aligned}
	\end{equation}
	where $[x]_{+}=\mathop{\max}\{0,x\}$.
	
	The motivation of GANs is to train the generator based on the output of the discriminator. Unlike the direct training objective of the classification task (minimizing cross-entropy loss), the objective of generator is indirect (with the help of the discriminator output). Hence, discriminator should provide a richer representation on the truth or false of samples compared to the classification task. More prior information and additional supervision tasks are urgent during the training process of the discriminator. Based on these, some regularization methods, such as {\textit{Data Augmentation and Preprocessing}}, {\textit{Consistency Regularization}}, and {\textit{Self-Supervision}} are proposed to improve the stability and generalizability \cite{than2021generalization} of discriminator.
	
	
	\subsection{Middle Level: The Perspective of "Fitting distribution"}
\begin{figure}
		\centering
		\includegraphics[scale=0.55]{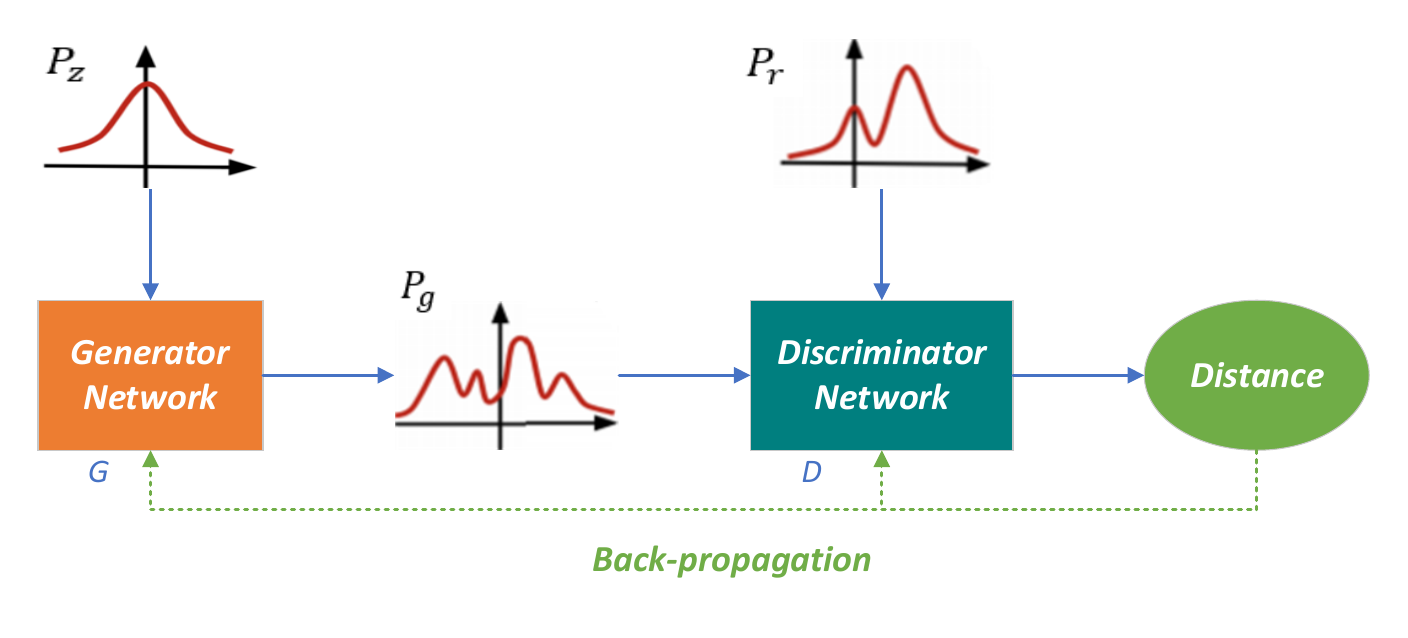}
		\caption{The framework of GANs. $P_z$ is a latent space distribution, $P_r$ and $P_g$  represent the real distribution and the generated distribution, respectively.}
		\label{FIG}
	\end{figure}

	At middle level, generator $G(z)$ is considered as a distribution mapping function that maps latent distribution $p_z(z)$ to generated distribution $P_g(x)$, and the discriminator $D(x)$ is a distribution distance that evaluates the distance between the target distribution $P_r(x)$ and the generated distribution $P_g(x)$ as illustrated in Figure \ref{FIG}. For the optimal discriminator, the generator $G(z)$ tries to minimize the distance between $P_r(x)$ and $P_g(x)$. For instance, generator of the vanilla GAN\footnote{Vanilla GAN, also known as standard GAN, is the first GAN model.} \cite{goodfellow2014generative} and $f$-GAN\footnote{\label{ft:5} $f$-GAN is a collective term for a type of GAN models whose discriminator minimizes $f$ divergence. $f$ divergence is the general form of KL divergence. It can be demonstrated as: $D_f(P||Q)=\int q(x)f\big(\frac{p(x)}{q(x)}\big)\rm{d}x$, where $f$ is a mapping function from non-negative real numbers to real numbers ($\mathbb{R}^*\rightarrow\mathbb{R}$) that satisfies: (1) $f(1)=0$. (2) $f$ is a convex function. To be more specific, KL divergence corresponds to $f(u)=u\log u$ and JS divergence corresponds to $f(u)=-\frac{u+1}{2}\log\frac{1+u}{2}+\frac{u}{2}\log u$. More details can be viewed in \cite{nowozin2016f} } \cite{nowozin2016f} are considered to minimize Jensen–Shannon (JS) divergence and $f$ divergence\textsuperscript{\ref{ft:5}}, respectively. When the conditions of LSGANs loss are set to $b-c=1$ and $b-a=2$, generator of the LSGAN considers the minimization of Pearson ${\chi}^2$ divergence. Furthermore, generators of the WGAN-div\footnote{Different from WGAN-div, WGAN \cite{arjovsky2017wasserstein} minimize Wasserstein distance, not Wasserstein divergence.} \cite{wu2018wasserstein} and GAN-QP \cite{su2018gan} consider the minimization of Wasserstein divergence and Quadratic divergence, respectively.
	
	Generator is a transportation map from $z$ to $p_g(x)$. In this section, we introduce the optimal transport and the optimal transport with regular term, which leads to the form of Wasserstein GANs with gradient penalty \cite{gulrajani2017improved} (WGAN-GP) and Wasserstein GANs with Lipschitz penalty \cite{petzka2017regularization} (WGAN-LP), respectively. Wasserstein distance is a popular and important distance in GANs and it corresponds to the optimal transport of the generator. To solve the dual problem of Wasserstein distance, Lipschitz continuity is introduced, which is the reason why gradient penalty and weight normalization techniques are proposed in the GANs training.

	The details of optimal transportation and optimal transportation with regular terms for WGAN and Lipschitz continuity can be found on the Section \ref{sect:A-1} of the Supplementary Online-only Material. It is pertinent to note that \textit{Gradient Penalty} and \textit{Gradient} \textit{Normalization} are two simple and effective ways to implement the Lipschitz continuity. Furthermore, \cite{miyato2018spectral} demonstrates that the spectral norm and the Lipschitz constant have the same meaning. Therefore, the spectral norm can be used to represent the Lipschitz constant. The Lipschitz continuity is achieved by normalizing the spectral norm of the weight, approximately. Hence, \textit{Weight Normalization} and \textit{Weight Regularization} can also be used to enable the Lipschitz continuity of the discriminator.

	\subsection{High Level: The Perspective of "Training dynamics"}
	GANs training is a two-player zero-sum game with a solution to Nash Equilibrium. At high level, we analyze the convergence of GANs by understanding the optimization process. Based on these, some regularization techniques are proposed to guide the GANs model to reach the theoretical equilibrium point leading to improvement in the effectiveness of GANs.
	
	Reconsidering the Eq (\ref{EQ:eqn1}) in Section 2, the training of GANs is achieved by solving a two-player zero-sum game via Simultaneous Gradient Descent (SimGD) \cite{goodfellow2014generative,arjovsky2017wasserstein}. The updates of the SimGD are given as:
	\begin{equation} 
		\label{eq:eqn27}
		\phi^{(k+1)}=\phi^{(k)}-h\nabla_\phi f(\phi^{(k)},\theta ^{(k)}),\quad
		\theta^{(k+1)}=\theta^{(k)}+h\nabla_\theta f(\phi^{(k)},\theta ^{(k)}).
	\end{equation}
	Assuming that the objectives of GANs are convex, many research studies discuss their global convergence characteristics \cite{nowozin2016f,yadav2017stabilizing}. However, due to the high non-convexity of deep networks, even a simple GAN does not satisfy the convexity assumption \cite{nie2019towards}. A recent study \cite{li2017limitations} shows that it is unrealistic to obtain approximate global convergence under the assumption of the optimal discriminator, so the community considers local convergence. It hopes that the trajectory of the dynamic system can enter a local convergence point with continuity iterations, that is, Nash equilibrium:
	\begin{equation} 
		\bar{\phi}=\mathop{\arg\max}_{\phi}-f(\phi,\bar{\theta}),\quad
		\bar{\theta}=\mathop{\arg\max}_{\theta}f(\bar{\phi},\theta).
		\label{EQ:eq24}
	\end{equation}
	If the point $(\bar{\phi},\bar{\theta})$ is called the local Nash-equilibrium, Eq  (\ref{EQ:eq24}) holds in a local neighborhood of  $(\bar{\phi},\bar{\theta})$.
	For this differentiable two-player zero-sum game, a vector is defined as below:
	\begin{equation}
		v(\phi,\theta)=
		\left(
		\begin{aligned}
			-\nabla_\phi f(\phi,\theta)\\
			\nabla_\theta f(\phi,\theta)
		\end{aligned}  
		\right).
	\end{equation}
	The Jacobian matrix is:
	\begin{equation}
		v^{'}(\phi,\theta)=
		\left(
		\begin{aligned}
			-&\nabla^2_{\phi,\phi} f(\phi,\theta)\quad-\nabla^2_{\phi,\theta} f(\phi,\theta)\\
			&\nabla^2_{\phi,\theta} f(\phi,\theta)\quad\nabla^2_{\theta,\theta} f(\phi,\theta)
		\end{aligned}  
		\right).
	\end{equation}
	
	According to the propositions on the Section \ref{sect:A-2} of the Supplementary Online-only Material, under the premise of asymptotic convergence, the local convergence of GAN is equivalent to the absolute value of all eigenvalues of the Jacobian matrix at the fixed point $(v(\bar{\phi},\bar{\theta})=0)$ being less than 1. To get this condition, \textit{Jacobian Regularization} \cite{mescheder2017numerics,nagarajan2017gradient,mescheder2018training,roth2017stabilizing} needs to be used. 
	\section{Regularization and Normalization of "Real \& Fake"}
	From the perspective of the "Real \& Fake", generator is counterfeiter designed to deceive the discriminator, while discriminator is police designed to distinguish between real and fake samples. The motivation of GANs is to train the generator based on the loss of the discriminator. {Compared to supervised classification tasks, discriminator formally needs to perform only bi-classification tasks, which is easy to implement. Therefore, discriminator is very easy to overfit.} Furthermore, unlike the direct training objective of the classification task (Minimizing cross-entropy loss), the objective of GANs training is indirect. Hence, only one-dimensional output of the discriminator does not provide a complete representation on truth or false of samples. Some studies have shown that the present discriminators contain some significant deficiencies in the frequency domain \cite{chen2020ssd} and attribute domain \cite{zhao2018bias}, which are evidence of the lacking discrimination for discriminators. Excessive shortage of discrimination makes the generator lack incentives from the discriminator to learn useful information of the data. {In addition to discriminator overfitting and lacking discrimination of discriminators, discriminator forgetting is another challenge for GANs.} To alleviate these situations, many regularization methods and additional supervision tasks have been proposed in the literature, which can be divided into three categories: \textit{Data Augmentation and Preprocessing}, \textit{Consistency Regularization}, and \textit{Self-supervision}. {All of them are based on data augmentation and are orthogonal to each other. As shown in Table \ref{table:SOTA }, The state-of-the-art GANs always adopt two or even all of the above regularization.}
	\subsection{Data Augmentation and Preprocessing}
	\begin{figure}
		\centering
		\includegraphics[scale=.35]{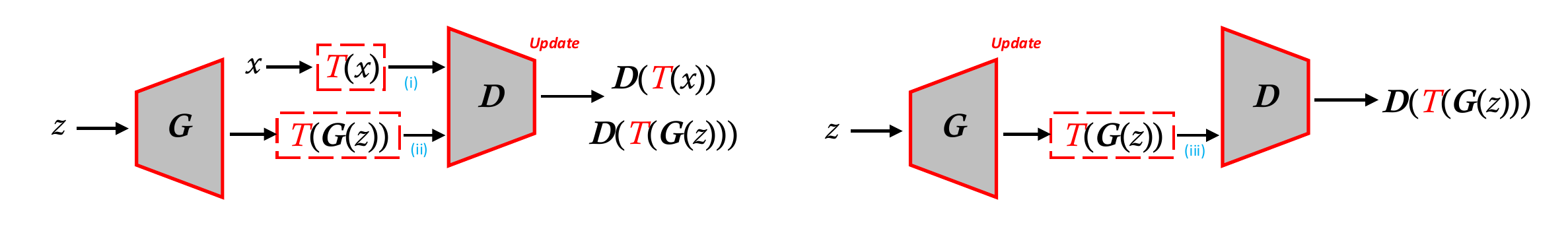}
		\caption{Framework of data augmentation and preprocessing for updating D (left) and G (right). (Coming from \cite{zhao2020differentiable})}
		\label{fig:data_augmentation_framework}
	\end{figure}
	
	Data Augmentation plays a significant role in deep learning algorithms. It increases the diversity of the training data naturally, thus reduces the overfitting in many computer vision and graphics applications \cite{krizhevsky2017imagenet,wan2013regularization}. Date augmentation adopts different data transformation techniques ({\color{red}$T$}) to increase the number of training samples. One type of  data transformation is spatial transformation of data, such as $zoomout$, $zoomin$, $translation$, $translationx$, $translationy$, $cutout$ \cite{devries2017improved}, $cutmix$ \cite{yun2019cutmix}; The other is
	visual transformation, such as $brightness$, $redness$, $greenness$, $blueness$, $mixup$ \cite{zhang2017mixup}. {Furthermore, recent study \cite{li2021high} is also attempting to use frequency transformation \cite{xia2022cscnet,wong2021view} to data augmentation. }

	Similarly, the performance of GANs heavily deteriorates given a limited amount of training data \cite{tseng2021regularizing}. {For instance, \cite{karras2020training} shows that Frechet Inception Distance (FID) starts to rise at some point during training and outputs of discriminator keep drifting apart during training, when training data is limited. More analysis can be found in the survey \cite{li2022comprehensive} on data-efficient GANs training.} However, recent studies \cite{zhang2019consistency, zhao2020differentiable,zhao2020image,karras2020training,tran2020towards} observe that augmenting only real images (Only applying $T$ to (i) in Figure \ref{fig:data_augmentation_framework}), only generated images (Only applying $T$ to (ii) in Figure \ref{fig:data_augmentation_framework}), and only discriminator (Both applying $T$ to (i) and (ii) in Figure \ref{fig:data_augmentation_framework}) do not help with GANs training. Naturally, one problem needs to be considered: whether the overfitting exists in GANs' training? Some studies \cite{zhao2020differentiable,karras2020training} demonstrate that, even with big dataset and state of the art models, the training of GANs suffers from severe overfitting. Furthermore, in case of small training data, overfitting occurs at an early stage in the training. Recently, some studies \cite{zhao2020differentiable,zhao2020image,karras2020training,tran2020towards,jiang2021deceive} on data augmentation for GANs training have been proposed. It is argued that the classical data augmentation approach could mislead the generator to learn the distribution of the augmented data, which could be different from that of the original data. To deal with this problem, these studies augment both real and fake samples and let gradients propagate through the augmented samples to G (Applying $T$ to (i), (ii), and (iii) in Figure \ref{fig:data_augmentation_framework}). By adding the data augmentation to all processes of GANs training, the performance of GANs has been significantly improved. {However, this "\textit{Augment All}" strategy may lead to the “leaking” of augmentations to the generated samples, which is highly undesirable. The experiments in \cite{karras2020training} demonstrate that as long as the probability of executing the augmentation remains below 0.8, leaks are unlikely to happen in practice.}
	
	Data augmentation in GANs has remarkable achievement. However, which augmentation is most beneficial for GANs training is still an open problem. Figure 2 in \cite{zhao2020image} shows the FID comparisons of BigGAN on CIFAR-10 dataset. For data augmentation (represented by ‘vanilla\_rf’), the operations in spatial augmentation such as $translation$, $zoomout$, and $zoomin$, are much more effective than the operations in visual augmentation, such as $brightness$, $colorness$ ($redness$ and $greenness$), and $mixup$. The results indicate that augmentation leads to spatial changes which improves GANs performance compared with cases where visual changes are induced. It is easy to understand that generated images are significantly lacking in detail information compared to the real images, and spatial augmentation improves the ability of the generator to fit detailed textures through spatial changes. {$InstanceNoise$, resulting in images out of the natural data manifold, cannot help with improving GANs performance. Apart form applying only a limited range of augmentations, some studies explore some strong data augmentations in GANs training. For instance, Jeong et al. \cite{jeong2021training} adopt contrastive learning to extract more useful information under stronger data augmentation beyond the existing yet limited practices. Combining adaptive strategies and 18 transformations (Both spatial and visual transformations) \cite{karras2020training}, even and $InstanceNoise$ only \cite{wang2022diffusion} can bring performance improvement over strong GANs baselines. {Furthermore, \cite{wang2022diffusion} is the first method to tackle the generative learning trilemma with denoising diffusion GANs.} Apart from these, Jiang et al. \cite{jiang2021deceive} also devise an adaptive strategy to control the strength of selecting generated images to augment real data, which can further boosts the performance of GANs.  Data augmentation is popular and significant in GANs training, whose achievements are attributed to improving discrimination, avoiding overfitting, and increasing the overlap \cite{zhao2020image,wang2022diffusion} between real and fake distributions.}

	
	{Different from data augmentation that increases the amount of the training data, data preprocessing only adopt prior knowledge and do some uniform data transformation before the network training. Data preprocessing is orthogonal to data augmentation, which can further enhance the performance combining the data augmentation.} Li et al. \cite{li2021high} indicate that high-frequency components between real and fake images are different, which is not conducive to the GANs training. They propose two preprocessing eliminating high-frequency differences in GANs training: High-Frequency Confusion (HFC) and High-Frequency Filter (HFF). These methods are applied in places (i), (ii), and (iii) in Figure \ref{fig:data_augmentation_framework} and improve the performance of GANs with a fraction of the cost. 
	
	In summary, both data augmentation and data preprocessing improve the performance of GANs with little cost. Data augmentation uses different transformations to improve discrimination and avoid disciminator overfitting. Furthermore, spatial augmentations achieve better performance than visual augmentations. More specifically, Zhao et al. \cite{zhao2020differentiable} demonstrate that hybrid augmentation with $Color + Translation + Cutout$  is especially effective and is widely used in other studies \cite{jeong2021training,anokhin2020image}.  Adaptive data augmentation (ADA) is the most popular method in GANs. Besides the data augmentation, data preprocessing is also a remarkable method.
	
	\begin{figure}
		\centering
		\includegraphics[scale=.3]{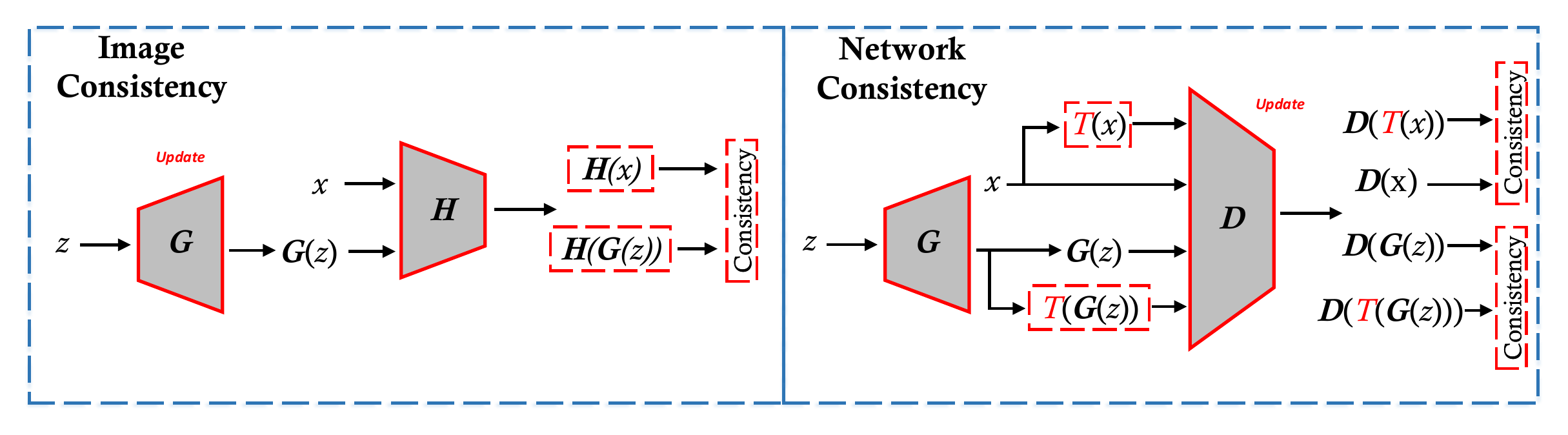}
		\caption{Overview of consistency regularization, where image consistency regularization updates the G (left) and network consistency regularization updates the D (right). $H$ is the feature mapping function and {\color{red}$T$} is different data transformation techniques.}
		\label{fig:consistency}
	\end{figure}
	
	\subsection{Consistency Regularization}
	In context of semi-supervised and unsupervised learning, consistency regularization has been widely used in \cite{xie2019unsupervised,sohn2020fixmatch,gao2019consistency,chen2020simple}. It is motivated by the fact that models should produce consistent predictions given input and their semantics-preserving augmentations, such as image rotating, and adversarial attacks. It is pertinent to note that the supervision of GANs training is weak. To increase the discrimination of discriminator, some consistency regularization techniques have also been used. Due to different goals, we divide these into two parts: \textit{Image Consistency} and \textit{Network Consistency} as demonstrated in Figure \ref{fig:consistency}.
	
	\subsubsection{Image Consistency}
	The purpose of GANs is to generate fake images similar to real ones. In GANs, the discriminator is generally used to distinguish real images and generated images. However, outputs of the discriminator with only one dimension hardly portray the authenticity of the image completely. To improve the representation of the discriminator, some studies extend the outputs of the discriminator, for example, relativistic discriminator \cite{jolicoeur2018relativistic,su2018training}, distribution discriminator \cite{xiangli2020real}, and cascading rejection \cite{shin2019simple}. Apart from this, some studies reduce the training difficulty of discriminators by introducing prior information. Regularizing the distance between the generated and real images with different measurements, namely image consistency, is the focus of this paper. The overview of it is demonstrated in left part of Figure \ref{fig:consistency}, where consistency regularization is used to update generator (G) and can be formulated as:
	\begin{equation}
		\mathcal{L}_C=\mathbb{E}_{x\sim p_r, z\sim p_z} C(H(x),H(G(z))),
		\label{EQ:image_consistency}
	\end{equation}
	where $H$ is the feature mapping function and $C$ is the consistency measurement function. Different image consistency regularization have different $H$ and $C$. For instance, Salimans et al. \cite{salimans2016improved} recommend that the generator is trained using a feature matching procedure. The objective is:
	\begin{equation}
		\mathcal{L}_C=||\mathbb{E}_{x\sim p_r} f(x)-\mathbb{E}_{z\sim p_z}f(G(z))||^2_2,
	\end{equation}
	where $f(x)$ denotes the intermediate layer of the discriminator. Similarly, the intermediate layer of another pre-trained classification model is an alternate option. The empirical results indicate that feature matching is indeed effective in situations where normal GAN becomes unstable. Unlike the above study which only uses mean feature matching to training generators, Mroueh et al. \cite{mroueh2017mcgan} propose McGAN, which trains both the generator and discriminator using the mean and covariance feature matching. The objective is:
	\begin{equation}
			\mathcal{L}_C=\mathcal{L}_\mu+\mathcal{L}_\sigma
			=||\mu(p_r)-\mu(G(p_z))||_q+||\sum\left(p_r\right)-\sum(p(G(z)))||_k,
	\end{equation}
	where $\mu(p_r)=\mathbb{E}_{x\sim p_r} f(x)$ and $\sum(p_r)=\mathbb{E}_{x\sim p_r} f(x)\cdot f(x)^\mathrm{T}$ represent the mean and the covariance of the feature layer $f(x)$, respectively. Apart from statistical differences, some studies \cite{durall2020watch,chen2020ssd} focus on the difference in frequency domain between the generated and real image. For instance, Durall et al. \cite{durall2020watch} find that the deep generative models based on up-convolution fail to reproduce spectral distributions leading to considerable differences in the spectral distributions between real images and generated images. Thus, the spectral regularization has been proposed as follows:
	\begin{equation}
		\mathcal{L}_C=\frac{1}{M/2-1}\sum_{i=0}^{M/2-1}AI^{real}_i\cdot\log(AI^{fake}_i)
		+(1-AI^{real}_i)\cdot\log(1-AI^{fake}_i),
	\end{equation}
	where $M$ is the image size and $AI$ is the spectral representation from the Fourier transform of the images. Corresponding to Eq (\ref{EQ:image_consistency}), $H$ and $C$ are implemented with $AI$ and cross-entropy, respectively.
	
	Contrary to this, the research study \cite{chen2020ssd} uses hard example mining to improve the discriminatory of the discriminator based on the difference between real and generated samples under different metrics. Although this paradigm is different from the paradigm of image consistency regularization, both cases are motivated by obtaining generated samples similar to real images under different distance measures, so we integrate them. Chen et al. \cite{chen2020ssd} consider both downsampling strategies: downsampling with anti-aliasing and downsampling without anti-aliasing, leads to high frequencies missing in the discriminator. High frequencies missing leads to high frequency deviation between real and generated images. To mitigate this issue, authors propose SSD-GAN, which introduces an additional spectral classifier to detect frequency spectrum discrepancy between real and generated images and integrate it into the discriminator of GANs. The overall realness of sample x is represented as:
	\begin{equation}
		D^{s s}(x)=\lambda D(x)+(1-\lambda) C(\phi(x)),
	\end{equation}
	where the enhanced discriminator $D^{s s}$ consists of two modules, a vanilla discriminator $D$ that measures the spatial realness, and a spectral classifier $C$. $\lambda$ is a hyperparameter that controls the relative importance of the spatial realness and the spectral realness. The adversarial loss of the framework can be written as:
	\begin{equation}
		\mathcal{L}_{D} =\mathbb{E}_{x \sim p_{\text {data }}(x)}\left[\log D^{s s}(x)\right] 
		+\mathbb{E}_{x \sim p_{g}(x)}\left[\log \left(1-D^{s s}(x)\right)\right].
	\end{equation}
	{karnewar et al. \cite{karnewar2020msg} introduce adversarial loss into the intermediate layer of the generator, which provides multiple and richer metrics for the training of generator.}
	
	\begin{table}
		\caption{The summary of the consistency regularization.}
		\label{table:consistency}
		\tiny
		\centering
		\begin{tabular}{c| c  }	
			\toprule
			\midrule
			Method& Consistency regularization term $L_C$	 \\
			\hline
			Mean regularization \cite{salimans2016improved}&$||\mathbb{E}_{x\sim p_r} f(x)-\mathbb{E}_{z\sim p_z}f(G(z))||_q$\\
			\hline
			Mean and Convariance regularization \cite{mroueh2017mcgan}&$||\mathbb{E}_{x\sim p_r} f(x)-\mathbb{E}_{z\sim p_z}f(G(z))||_q+||\mathbb{E}_{x\sim p_r} f(x)\cdot f(x)^\mathrm{T}-\mathbb{E}_{z\sim p_z}f(G(z))\cdot f(G(z))^\mathrm{T}||_k$\\
			\hline
			Spectral regularization \cite{durall2020watch}&$\mathcal{L}_C=\frac{1}{M/2-1}\sum_{i=0}^{M/2-1}AI^{real}_i\cdot\log(AI^{fake}_i)+(1-AI^{real}_i)\cdot\log(1-AI^{fake}_i)$\\
			\hline
			
			CR-GAN \cite{zhang2019consistency,ohkawa2020augmented}&$\mathbb{E}_{x\sim p_r}|| D(x)-D(T(x))||_2$\\
			\hline
			bCR-GAN \cite{zhao2020improved}&$\lambda_{real}\mathbb{E}_{x\sim p_r}|| D(x)-D(T(x))||_2+\lambda_{fake}\mathbb{E}_{z\sim p_z}||D(G(z))-D(T(G(z)))||_2$\\
			\hline
			zCR-GAN\cite{zhao2020improved}&$\lambda_{dis}\mathbb{E}_{z\sim p_z}||D(G(z))-D(G(T(z)))||_2-\lambda_{gen}\mathbb{E}_{z\sim p_z}||G(z)-G(T(z))||_2$\\
			\bottomrule
			
		\end{tabular}
	\end{table}
	
	The summary of the image consistency regularization is given in Table \ref{table:consistency}. In summary, image consistency considers that the real images and the generated images are similar not only in the output of discriminator, but also in statistical information and frequency domain. The analysis of biases between real and generated images using different metrics will be an interesting future research direction.
	\subsubsection{Network Consistency}
	Network consistency regularization can be regarded as Lipschitz continuity on semantics-preserving transformation. Specifically, we hope discriminator is insensitive to semantics-preserving transformation, which drives the discriminator to pay more attention to the authenticity of the images. For example, in the image domain, the reality of images should not change if we flip the image horizontally or translate the image by a few pixels. To resolve this, Zhang et al. \cite{zhang2019consistency} propose the Consistency Regularization GAN (CR-GAN) that uses the consistency regularization on the discriminator during GANs training:
	\begin{equation}
		\mathcal{L}_C=\mathbb{E}_{x\sim p_r}|| D(x)-D(T(x))||_2,
	\end{equation}
	where $T(x)$ represents a transformation (shift, flip, cutout, etc.) of images. One key problem with the CR-GAN is that the discriminator might occur the 'mistakenly believe'. 'mistakenly believe' considers that the transformations are actual features of the target dataset, due to only applying these transformations on real images. This phenomenon is not easy to notice for certain types of transformations (e.g. image shifting and flipping). However, some types of transformations, such as cutout transformations, contain visual artifacts not belonging to real images, which effects greatly limits the choice of advanced transformations we could use. To address this issue, Zhao et al. \cite{zhao2020improved} propose Balanced Consistency Regularization (bCR-GAN) that uses regulation with respect to both real and fake images and balances the training of discriminator between real images and fake images by $\lambda_{real}$ and $\lambda_{fake}$:
	\begin{equation}
		\mathcal{L}_C=\lambda_{real}\mathbb{E}_{x\sim p_r}|| D(x)-D(T(x))||_2
		+\lambda_{fake}\mathbb{E}_{z\sim p_z}||D(G(z))-D(T(G(z)))||_2.
	\end{equation}
	The overview of bCR is demonstrated in right part of Figure \ref{fig:consistency}.
	
	Contrary to the methods which focus on consistency regularization with respect to transformations in image space, Zhao et al. \cite{zhao2020improved} also propose Latent Consistency Regularization (zCR) that considers the consistency regularization on transformations in latent space. Authors expect that output of the discriminator ought not to change much with respect to the small enough perturbation $\Delta z$ and modify the discriminator loss by enforcing:
	\begin{equation}
		\mathcal{L}^D_C=\lambda_{dis}\mathbb{E}_{z\sim p_z}||D(G(z))-D(G(T(z)))||_2,
	\end{equation}
	where $T(z)$ represents the added small perturbation noise. However, if only this loss is added into the GAN loss, mode collapse can easily appear in the training of generators. To avoid this, an inverse gradient penalty (we will describe it in section 6.2) is added to modify the loss function for generator. Hence, we modify the generator loss by enforcing:
	\begin{equation}
		\mathcal{L}^D_C=-\lambda_{gen}\mathbb{E}_{z\sim p_z}||G(z)-G(T(z))||_2.
	\end{equation}
	Naturally, putting both bCR and zCR together, Improved Consistency Regularization (ICR) is also proposed by Zhao et al. \cite{zhao2020improved}. In addition, there are some applications where cyclic consistency regularization is used for unpaired image-to-image translation \cite{ohkawa2020augmented}. The summary of network consistency regularization is given in Table \ref{table:consistency}. 
	
	In summary, network consistency considers the networks, especially the discriminator, to be insensitive to semantics-preserving transformation ($T$). The results in \cite{zhang2019consistency} demonstrate that random shift and flip is the best way to perform image transformation on the CIFAR-10 dataset. Furthermore, the FID results with CR, bCR, zCR, and ICR (where transformation is flipping horizontally and shifting by multiple pixels) as presented in \cite{zhao2020improved} are shown in Table \ref{table:consistency_results}. The results demonstrate that network consistency regularization can significantly improve the performance of GANs. However, which transformation is best for consistency regularization, is a question. Zhao et al \cite{zhao2020image} compare the effect of different data transformation techniques (mentioned in Section 5.1) on bCR. Figure 2 in \cite{zhao2020image} shows the FID results of BigGAN adding bCR (represented by 'bcr') on CIFAR-10 dataset. From the results, the best BigGAN FID 8.65 is with transformation technology $translation$ of strength $\lambda=0.4$, outperforming the corresponding FID 10.54 reported in Zhao et al. \cite{zhao2020improved}. Moreover, spatial transforms, which retain the major content while introducing spatial variances, can substantially improve GANs performance together with bCR. While visual transforms, which retain the spatial variances, can not further improve the performance of GANs compared with data augmentation only. Furthermore, bCR with stronger transformation (larger value of $\lambda_{aug}$) does not improve the performance of GANs, the optimal value of $\lambda_{aug}$ is uncertain for different data transformations. 
	\begin{table}
		\caption{FID scores for class conditional image generation of the network consistency regularization (Data come from \cite{zhao2020improved}).}
		\label{table:consistency_results}
		\centering
		\begin{tabular}{ccc}
			\toprule
			\midrule
			Models & CIFAR-10 & ImageNet \\
			\hline SNGAN & 17.50 & 27.62 \\
			BigGAN & 14.73 & 8.73 \\
			CR-BigGAN & 11.48 & 6.66 \\
			bCR-BigGAN & 10.54 & 6.24 \\
			zCR-BigGAN & 10.19 & 5.87 \\
			ICR-BigGAN & $\mathbf{9 . 2 1}$ & $\mathbf{5 . 3 8}$ \\
			\hline
		\end{tabular}
	\end{table}
	
	\subsection{Self-Supervision}
	Self-supervised learning aims to learn representations from the data itself without explicit manual supervision. Recently, some self-supervised studies \cite{he2020momentum,hjelm2018learning,chen2020simple} provide competitive results on ImageNet classification and the representations learned from which transfer well to downstream tasks. Self-supervised learning outperforms its supervised pre-training counterpart in many tasks, such as detection and segmentation, sometimes surpassing it by large margins. This suggests that self-supervised learning obtains more representational features and significantly improve the representation of networks. Based on this, self-supervised learning is introduced into the training of GANs, and we divide them into two categories according to different self-supervision tasks: \textit{Predictive Self-supervised Learning} and \textit{Contrastive Self-supervised Learning}.
	
	\subsubsection{Predictive Self-Supervised Learning (PSS)}
\begin{figure}
		\centering
		\includegraphics[scale=.4]{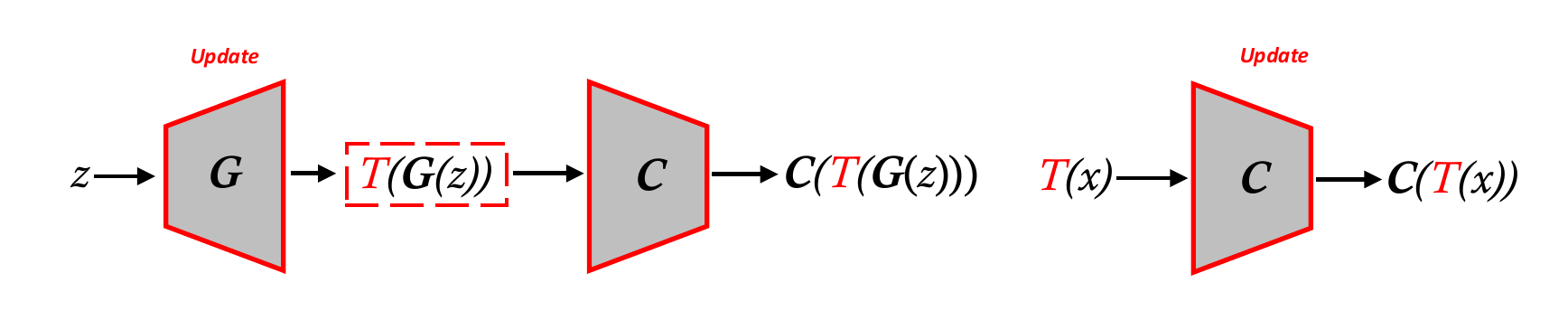}
		\caption{Overview of the predict self supervised learning of GANs, where $C$ performs the predict classification task and shares the weights with the discriminator except for the last layer, {\color{red}$T$} is  different data transformation techniques. Furthermore, the self-supervised task of generated images is used to update the generator (left) and the self-supervised task of real images is used to update the classification (right).}
		\label{fig:predict_self_supervised}
	\end{figure}

	Predictive self-supervised learning is a popular method to improve the representation of neural networks by introducing additional supervised tasks, such as context  prediction \cite{doersch2015unsupervised} and rotations prediction \cite{gidaris2018unsupervised,lee2019rethinking,kolesnikov2019revisiting,zhai2019s4l}. Predictive self-supervised learning is introduced into GANs by Chen et al. \cite{chen2019self} to avoid discriminator forgetting. Discriminator forgetting means that the discriminator does not remember all tasks at the same time during the training process. For example, learning varying levels of detail, structure, and texture, which causes the discriminator to fail to get a comprehensive representation of the current images. "If the outcome is your focus, then it’s easy to look for shortcuts. And ultimately shortcuts keep you from seeing the truth, it drains your spark for life. What matters most is your will to seek the truth despite the outcome."\footnote{Come from "JoJo's Bizarre Adventure:Golden Wind" -Araki Hirohiko.}. The same is true for GANs, which are only driven by the loss of discriminator, which is easy to distinguish between real images and generated images through shortcuts, instead of the texture and structural features we need. Predictive self-supervised learning solves this problem by introducing new generalization tasks, which also helps to prevent overfitting. The overview of the predictive self-supervised learning of GANs are demonstrated on Figure \ref{fig:predict_self_supervised}. Depending on the data transformation function $T$, we can design different self-supervised tasks. 
	
	Chen et al. \cite{chen2019self} introduced the predictive self-supervision in GANs training. Authors adopt the rotation prediction as the expanding task to prevent the discriminator from forgetting. Besides, plenty of other prediction tasks have also been proposed to improve the discrimination. Huang et al. \cite{huang2020fx} exploit the feature exchange to make the discriminator learn the proper feature structure of natural images. Baykal et al. \cite{baykal2020deshufflegan} introduce a reshuffling task to randomly arrange the structural blocks of the images, thus helping the discriminator increase its expressive capacity for spatial structure and realistic appearance. Contrary to the methods for designing tasks at the image or feature level, Patel et al. \cite{patel2021lt} propose a self-supervised task with latent transformation detection, which identifies whether the latent transformation applied in the given pair is the same as that of the other pair. All above methods have designed different self-supervised tasks, and their loss functions can be formulated as:
	\begin{equation}
		\begin{split}
			\mathcal{L}_{D,C}=-\lambda_{r}\mathbb{E}_{x\sim p^T_r}\mathbb{E}_{T_k\sim {T}}\log\big( C_k(x)\big) \quad \text { for } k=1, \ldots, K,\\
			\mathcal{L}_G=-\lambda_{g}\mathbb{E}_{x\sim p^T_g}\mathbb{E}_{T_k\sim {T}}\log\big( C_k(x)\big) \quad \text { for } k=1, \ldots, K,
		\end{split}
	\end{equation}
	where ${T}$ represents the different types of image transfer, such as rotation and reshuffling. Furthermore, $T_k$ represents different forms of the transfer ${T}$, such as $0^\circ,90^\circ,180^\circ, 270^\circ$ for rotation task, $K$ is the number of transformed forms, $C_k$ is the k-th output of the classifier $C$ that shares parameters with discriminator except for two different heads, $P^T_r$ and $P^T_g$ are the transformed distributions of real and generated images, respectively. For rotation conversion task \cite{chen2019self}, $K=4$, and the classifier $C$ predicts the rotation angle; For feature exchange task \cite{huang2020fx}, $K=2$, and the classifier $C$ predicts whether the swap has occurred; For block reshuffling task \cite{baykal2020deshufflegan}, the image is divided into $9$ blocks and the number of the permutations is $9!$, which is unnecessarily huge. Thirty different permutations are selected in terms of the Hamming distances between the permutations in \cite{carlucci2019domain}. As a result, $K$ is set to 30, and classifier $C$ predicts the Hamming distances of different permutations; For the latent transformation task, $K=2$, and the classifier $C$ predicts whether the transformations parameterized by the same $\epsilon$ or different. {Besides, some study \cite{liu2020towards} introduces the autoencoder task, making discriminator reconstruct the input. }
	\begin{table*}
		\renewcommand\arraystretch{1}
		\caption{The summary of the Self-supervision}
		\label{table:self-supervised}
		\setlength{\tabcolsep}{0.5mm}
		\centering
		\begin{tabular}{c| c |c }	
			\toprule
			\midrule
			\tiny Method&\tiny Description&\tiny Types\\
			\hline
			\tiny Rotation Prediction \cite{chen2019self, tran2019self}&\tiny Predicting the angle of rotation ($0^\circ,90^\circ,180^\circ, 270^\circ$)&\tiny PSS\\
			\hline
			\tiny Feature Exchange Detection \cite{huang2020fx}&\tiny Predicting if some exchanges have occurred at the feature level (yes or not)&\tiny PSS\\
			\hline
			\tiny Block Reshuffling Prediction \cite{baykal2020deshufflegan}&\tiny Predicting the Hamming distances of different reshuffling in image level (Total 30 categories)&\tiny PSS\\
			\hline 
			\tiny Latent Transformation Detection \cite{patel2021lt}&\tiny Predicting if some exchanges have occurred at the latent space level (yes or not)&\tiny PSS\\
			\hline
			\tiny Autoencoder \cite{liu2020towards}&\tiny Reconstruct the input of the discriminator&\tiny PSS\\
			\hline
			\tiny InfoMax-GAN \cite{lee2020infomax}&\tiny \makecell[c]{Positive pairs: Global and local features of an image (both real and fake images)\\ Negative pairs: Global and local features of different images (both real fake images)}&\tiny CSS\\
			\hline 
			\tiny Cntr-GAN \cite{zhao2020image}&\tiny \makecell[c]{Positive pairs: Two different data transformations of the same image (both real and fake images).\\ Negative pairs: Otherwise}&\tiny CSS\\
			\hline
			\tiny	ContraD \cite{jeong2021training}&\tiny \makecell[c]{Positive pairs: Two different data transformations of the same image (real images only) + Two fake images\\ Negative pairs: Otherwise}&\tiny CSS\\
			\hline
			\tiny	InsGen \cite{yang2021data}&\tiny \makecell[c]{Positive pairs: Two different data transformations (additional latent transformations for fake image)\\ of the same image (both real and fake images). Negative pairs: Otherwise.}&\tiny CSS\\
			\hline
			\tiny	FakeCLR \cite{li2022fakeclr}&\tiny \makecell[c]{Positive pairs: Two different data transformations and additional latent transformations for fake image.\\ Negative pairs: Otherwise.}&\tiny CSS\\
			\bottomrule
			
		\end{tabular}
	\end{table*}
	
	The above methods design different kinds of self-supervised prediction tasks and participate in the training of the discriminator or generator, independently, having “loophole” that, during generator learning, $G$ could exploit to minimize $\mathcal{L}_G$ without truly learning the data distribution. To address this issue, Ngoc-TrungTran et al. \cite{tran2019self} introduce true or false judgment along with self-supervised prediction. The number of classification is $K+1$, while the loss function can be expressed as:
	\begin{equation}
		\begin{aligned}
			\mathcal{L}_{D,C}=&-\lambda_{r}\Bigg(\mathbb{E}_{x\sim p^T_r}\mathbb{E}_{T_k\sim {T}}\log\big( C_k(x)\big)
			+\mathbb{E}_{x\sim p^T_g}\mathbb{E}_{T_k\sim {T}}\log\big( C_{K+1}(x)\big)\Bigg)\quad \text { for } k=1, \ldots, K,\\
			\mathcal{L}_G=&-\lambda_{g}\Bigg(\mathbb{E}_{x\sim p^T_g}\mathbb{E}_{T_k\sim {T}}\log\big( C_k(x)\big)
			-\mathbb{E}_{x\sim p^T_g}\mathbb{E}_{T_k\sim {T}}\log\big( C_{K+1}(x)\big)\Bigg)\quad \text { for } k=1, \ldots, K,
		\end{aligned}
	\end{equation}
	where $C_k$ is a classifier that predicts the rotation angles and $C_{K+1}$ is a classifier that predicts the truth of the images. The new self-supervised rotation-based GANs use the multi-class minimax game to avoid the mode collapse, which is better than the original predictive self-supervised paradigm.
	
	In summary, predictive self-supervised learning improves the discrimination by designing different self-supervised prediction tasks, among them, rotation prediction \cite{chen2019self} is widely used (\cite{zhao2020differentiable,zhao2020image}) for its simplicity and practicality. The summary of different methods is illustrated in Table \ref{table:self-supervised}. {All of these methods are juxtaposed with each other. However, few studies have used multiple self-supervised tasks simultaneously, and the use of multiple self-supervised tasks to improve GANs training is still an open problem.}
	
	\subsubsection{Contrastive Self-Supervised Learning (CSS)}
\begin{figure}
		\centering
		\includegraphics[scale=.3]{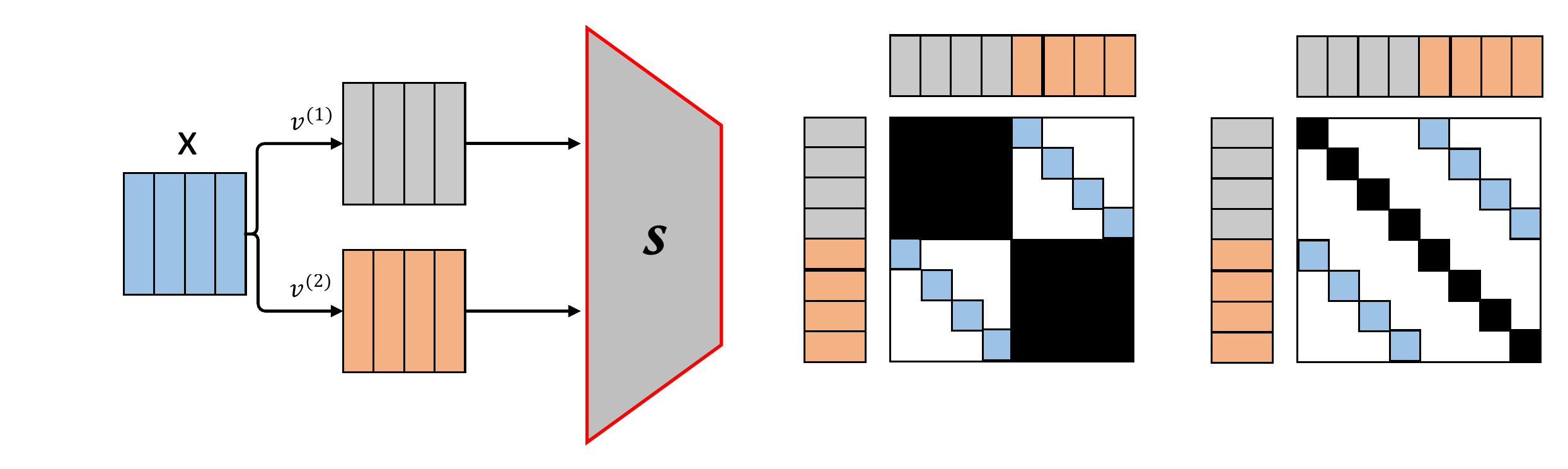}
		\caption{Overview of the contrastive self supervised learning, where x is real of fake images, $v^{(1)}$ and $v^{(2)}$ are different views of the image x, s is the score function (Usually discriminator in GANs) that measures the similarity, the square on the middle part is the label of InfoNCE (blue, white, and black squares are labels with 1, 0, and undefined respectively.), and the square on the right part is the label of SimCLR. Obviously, SimCLR defines more negative pairs to improve the sample utilization.}
		\label{fig:contrastive_self_supervised}
	\end{figure}

	Contrastive self-supervised Learning \cite{he2020momentum,hjelm2018learning,chen2020simple}, as the name implies, learn representations by contrasting positive and negative examples. These techniques have resulted in empirical success in computer vision tasks with unsupervised contrastive pre-training. A handful number of studies demonstrate that self-supervised learning outperforms its supervised pre-training counterpart in many tasks, which indicates contrastive self-supervised learning leads to more expressive features. Considering two views $(v^{(1)}$ and $v^{(2)})$, contrastive self-supervised learning aims to identify whether two views are dependent or not. More specifically, it means to maximize the mutual information of positive pairs. To this end, Oord et al. \cite{oord2018representation} propose to minimize InfoNCE loss, which turns out to maximize a lower bound of mutual information. The InfoNCE loss is defined by:
	\begin{equation}
		L_{\mathrm{NCE}}\left({v}_{i}^{(1)} ; {v}^{(2)}, s\right):=-\log \frac{\exp \left(s\left({v}_{i}^{(1)}, {v}_{i}^{(2)}\right)\right)}{\sum_{j=1}^{K} \exp \left(s\left({v}_{i}^{(1)}, {v}_{j}^{(2)}\right)\right)},
	\end{equation}
	where $s(,)$ is the score function that measure the similarity, positive pairs ($v_i^{(1)}$ and $v_i^{(2)}$) are different views of the same sample, and negative pairs ($v_i^{(1)}$ and $v_j^{(2)}$, $(i\neq j)$) are different views of different samples. InfoNCE loss is the cornerstone of contrastive self-supervised learning as depicted in Figure \ref{fig:contrastive_self_supervised}. 
	
	Many advanced self-supervised methods are implemented by modifying the views of images $v^{(1)}$, $v^{(2)}$ and score function $s(,)$. Specifically, Deep InfoMAX \cite{hjelm2019learning} maximizes the mutual information between local and global features, that is, image $x$ passes through the encoder $E_{\psi}=f_{\psi}\circ C_{\psi}$, producing local feature map $C_{\psi}(x)$ and global feature vector $E_{\psi}(x)$. To maximize the lower bound of the InfoMax: $\mathcal{I}\big(C_{\psi}(x),E_{\psi}(x)\big)$, the theoretical InfoMAX loss has been defined as:
	\begin{equation}
		\begin{aligned}
			L_{InfoMAX}(X)=&-\mathbb{E}_{x\in X}\mathbb{E}_{i\in\mathcal{A}}\big[
			\log \frac{\exp \left(g_{\theta, \omega}\left(C_{\psi}^{(i)}(x), E_{\psi}(x)\right)\right)}{\sum_{\left(x^{\prime}, i\right) \in X \times \mathcal{A}} \exp \left(g_{\theta, \omega}\left(C_{\psi}^{(i)}\left(x^{\prime}\right), E_{\psi}(x)\right)\right)}\big],\\
			g_{\theta, \omega}&\left(C_{\psi}^{(i)}(x), E_{\psi}(x)\right)=\phi_{\theta}\left(C_{\psi}^{(i)}(x)\right)^{T} \phi_{\omega}\left(E_{\psi}(x)\right),
		\end{aligned}
	\end{equation}
	where $X=\{x_1,\cdots,x_N\}$ is a set of random images and $\mathcal{A}=\{0,1,\cdots,M^2-1\}$ represents indices of a $M\times M$ spatial sized local feature map. Based on this, positive sample pairs are $C^{(i)}_{\phi}(x)$ and $E_{\phi}(x)$, and negative sample pairs are $C^{(i)}_{\phi}(x^{'})$ and $E_{\phi}(x)$), where $x^{'}$ is a different image from $x$. SimCLR \cite{chen2020simple} is another popular contrast learning framework, which applies two independent transformations, namely $t_1$ and $t_2$, to obtain the different views $v^{(1)},v^{(2)}=t_1(x),t_2(x)$. The loss function of SimCLR is defined as: 
	
	\begin{equation}
		\begin{aligned}
			L_{\mathrm{SimCLR}}\left({v}^{(1)}, {v}^{(2)}\right)=\frac{1}{N} \sum_{i=1}^{N}\left(L_{\mathrm{NCE}}\left({v}_{i}^{(1)} ;\left[{v}^{(2)} ; {v}_{-i}^{(1)}\right], s_{\mathrm{SimCLR}}\right)\right),
		\end{aligned}
	\end{equation}
	where ${v}_{-i}:={v} \backslash\left\{{v}_{i}\right\}$ and $s_{\mathrm{SimCLR}}$ is defined as:
	\begin{equation}
		s_{\text {SimCLR }}\left({v}^{(1)}, {v}^{(2)} ; f, h\right)=\frac{h\left(f\left({v}^{(1)}\right)\right) \cdot h\left(f\left({v}^{(2)}\right)\right)}{\tau \cdot\left\|h\left(f\left({v}^{(1)}\right)\right)\right\|_{2}\left\|h\left(f\left({v}^{(2)}\right)\right)\right\|_{2}}.
	\end{equation}
	As shown in Figure \ref{fig:contrastive_self_supervised}, SimCLR defines more negative pairs to improve the sample utilization compared to InfoNCE. {However, SimCLR \cite{chen2020simple} needs a large batch size to  obtain some sufficiently rich negative samples (In \cite{chen2020simple}, batch size is set to 4096). To alleviate the attachment of SimCLR to large batch size, MoCo \cite{he2020momentum} introduce a negative queue to store and update negative samples.}

	The self-supervised methods mentioned above are also widely applied to the training of GANs. 
	Inspired by Deep InfoMax \cite{hjelm2019learning}, Lee et al. \cite{lee2020infomax} propose InfoMax-GAN maximizing the mutual information between local and global features of real and fake images. The regularization of discriminator is expressed as:
	\begin{equation}
		\mathcal{L}_\mathrm{InfoMax-GAN}=\lambda_d \{L_{\mathrm{InfoMAX}}(x_{r})+L_{\mathrm{InfoMAX}}(x_{f}) \}.
	\end{equation}
	where $x_r$ and $x_f$ represent sets of real and fake images, respectively.
	
	Inspired by SimCLR \cite{chen2020simple}, some studies \cite{jeong2021training,zhao2020image} introduce different data transformation techniques to create positive and negative pairs during GANs training. Zhao et al \cite{zhao2020image} propose Cntr-GAN, where SimCLR loss is used to regularize the discriminator on two random augmented copies of both real and fake images. The regularization of discriminator for transformation $T$ is:
	\begin{equation}
		\mathcal{L}_{\mathrm{Cntr-GAN}}=\lambda_d \{L_{\mathrm{SimCLR}}(x_{r}, T(x_{r}))+L_{\mathrm{SimCLR}}(x_{f}, T(x_{f})) \}.
		\label{Eq:Cntr-GAN}
	\end{equation}
	They also compare the effect of different data transformation techniques (mentioned in Section 5.1) on Cntr-GAN. Figure 6 in \cite{zhao2020image} shows the FID results of BigGAN adding SimCLR loss on CIFAR-10 dataset. The results illustrate that spatial transformations still work better than visual transformations and the best FID of 11.87 is achieved by applying adjusted SimCLR transformations with the cropping/resizing strength of 0.3. Although, regularization of auxiliary SimCLR loss improves GAN training, but does not outperform existing methods based on simple data augmentations, $e.g.$, bCR (demonstrated on Figure 2 in \cite{zhao2020image}).

	To improve the efficiency of contrastive learning, Jeong et al. \cite{jeong2021training} propose Contrastive Discriminator (ContraD), a way of training discriminators of GANs using improved SimCLR. Different from Cntr-GAN with SimCLR loss on both real and generated images, ContraD uses the SimCLR loss on the real images and the supervised contrastive loss on the generated images. Supervised contrastive loss adopts the contrastive between real and generated images, required as a GAN discriminator. More concretely, for two views $v^{(1)},v^{(2)}=t_1(x),t_2(x)$ with ${t}_{1}, {t}_{2} \sim {T}$, the loss of real images are:
	\begin{equation}
		L_{\text {con }}^{+}\left(D, h_{{r}}\right)=L_{\mathrm{SimCLR}}\left({v}_{{r}}^{(1)}, {v}_{{r}}^{(2)} ; D, h_{{r}}\right),
	\end{equation}
	where $h_r$ is a projection head for this loss. However, the loss for generated images, an extended version of contrastive loss to support supervised learning by allowing more than one view to be positive. More concretely, they assume all the views from fake samples have the same label against those from real samples. Formally, for each $v_i^{(1)}$, the positive views are represented by $V_i^{(2)}$ that is a subset of $v^{(2)}$. The supervised contrastive loss is defined by:
	
	\begin{equation}
		L_{\text {SupCon }}\left({v}_{i}^{(1)}, {v}^{(2)}, V_{i+}^{(2)}\right)=
		-\frac{1}{\left|V_{i+}^{(2)}\right|} \sum_{{v}_{i+}^{(2)} \in V_{i+}^{(2)}} \log \frac{\exp \left(s_{{SimCLR}}\left({v}_{i}^{(1)}, {v}_{i+}^{(2)}\right)\right)}{\sum_{j} \exp \left(s_{{SimCLR}}\left({v}_{i}^{(1)}, {v}_{j}^{(2)}\right)\right)}.
	\end{equation}
	Using the notation,the ContraD loss for fake samples are:
	\begin{equation}
		L_{\text {con }}^{-}\left(D, h_{f}\right)=
		\frac{1}{N} \sum_{i=1}^{N} L_{\text {SupCon }}\left({v}_{f, i},\left[{v}_{f,-i} ; {v}_{{r}}^{(1)} ; {v}_{{r}}^{(2)}\right],\left[{v}_{f,-i}\right] ; D, h_{{f}}\right),
	\end{equation}
	where $v_f= t_3(G(z))$ is a random view of fake samples ($t_3\sim T$), and $v_{-i}=v\backslash \{v_i\}$ is subset of $v$ that does not contain $v_i$. It is pertinent to note that authors also use an independent projection header $h_f$ in this loss instead of $h_r$ in $L_{\text {con }}^{+}\left(D, h_{{r}}\right)$. {The adopted supervised contrastive learning on the fake images introduce the real/fake information into contrastive learning, which improves the efficiency of contrastive learning, thus improve the discrimination of the discriminator.}
	
	To sum up, ContraD learns its contrastive representation by minimizing the following regularization loss:
	\begin{equation}
		L_{\text {con }}\left(D, h_{{r}}, h_{{f}}\right)=L_{\text {con }}^{+}\left(D, h_{{r}}\right)+\lambda_{\text {con }} L_{{con}}^{-}\left(D, h_{{f}}\right).
	\end{equation}
	The experimental results show that ContraD consistently improves the performance of GANs compared to other methods, such as Cntr-GAN, DiffAug, bCR, and CR. However, ContraD with different data transformations is not discussed further. 
	
	{The achievement of above SimCLR-based contrastive learning methods depends on the sufficiently large batch size. However, large-scale GANs training often has only a small batch size for limited computational resources. Therefore, MoCo-based contrastive learning method, namely InsGen \cite{yang2021data}, has been introduced into GANs training. InsGen follows the MoCo-v2 \cite{chen2020improved} to store the various negative samples with an extra queue. Furthermore, it also introduces a latent space augmentation for fake images. Combining with ADA and MoCo-based contrastive learning, InsGen \cite{yang2021data} has achieved state-of-the-art performance on a variety of datasets and training settings. Recently, Li et al. \cite{li2022fakeclr} identify that only latent space augmentation for fake images brings the major performance improvement and contrastive learning in real images causes performance drop on limited data generation (DE-GANs\cite{li2022comprehensive}). Based on this, they propose FakeCLR, which only applies contrastive learning on perturbed fake samples and devises three related training techniques. The experimental results manifest the new state of the arts in both few-shot generation and limited-data generation.}
	
	In summary, contrastive self-supervised learning designs different positive and negative pairs and maximizes the mutual information of positive pairs according to the InfoNCE loss. Different from classification and segmentation tasks, two types of samples (real and fake images) exist for generating adversarial networks, which add more possibilities to the definition of positive and negative pairs. In the future, score-based contrastive lea  rning may be proposed during the training of GANs. The summary of contrastive self-supervised regularization techniques of GANs is given in Table \ref{table:self-supervised}.
	
	\subsection{Summary}
	{According to the perspective of \textbf{"Real $\&$ Fake"}, many regularization and normalization technologies inspired from supervised learning have been proposed to GANs training. The key point of them is improving the representation and generalizability of the discriminator. \textit{Data Augmentation and Preprocessing} is a basic operation containing many types such as spatial augmentation, visual augmentation, frequency augmentation, and noise augmentation. Among them, combining adaptive strategies and all augmentation \cite{karras2020training} has achieved the most remarkable achievement and has been employed as default operations in most GANs training. \textit{Consistency Regularization} and \textit{Self-supervision} are designed additional tasks based on data augmentation, which further improve the efficiency of data augmentation and extract more useful information under stronger data augmentation beyond the existing yet limited practices. Currently, combining contrastive self-supervised learning with adaptive data augmentation \cite{yang2021data,li2022fakeclr} has achieved state of the art in GANs training.}
	
	\section{Regularization and Normalization of "Fitting distribution" }
	From the perspective of "Fitting distribution", generator is considered as a distribution mapping function and the optimal discriminator is considered to be the distribution divergence. Wasserstein distance is a popular and important in GANs, and it corresponds to the optimal transport of the generator. To solve the dual problem of Wasserstein distance, Lipschitz continuity is introduced into the GANs training. The Wasserstein-based GANs (WGAN and WGAN-GP) have achieved remarkable results during the training. However, some studies \cite{stanczuk2021wasserstein,fedus2017many,kodali2017convergence} suggest that the success of WGAN-GP is not due to the Wasserstein distance and the Lipschitz constraint of discriminator may improve the performance and stability of GANs training regardless of the statistical distance used as a loss function. Therefore, the Lipschitz continuity of discriminator is an essential condition during GANs training. {Weight clipping \cite{arjovsky2017wasserstein} is a simple and the first solution to enforce a Lipschitz constraint, which clamps the weights of discriminator to a fixed box after each gradient update.} Furthermore, {gradient penalty}, {weight normalization}, and {weight regularization} are widely applied in GANs training for fulfilling Lipschitz continuity as summarized in subsequent subsections.
	\subsection{Gradient Penalty}
	
	Gradient penalty is a simple and direct way to fulfill Lipschitz continuity. Specifically,  K-Lipschitz continuity of the function $f$ can be accessed by $\mathop{\min}\mathbb{E}_{\hat{x}\sim\pi}(||\nabla f(\hat{x})||_2-{\rm K})^2$. According to the optimal transport theory mentioned on the Section \ref{sect:A-1} of the Supplementary Online-only Material, gradient penalty can be used for the approximation of $W_c(\mu,\upsilon)$ in WGANs, named WGAN-GP \cite{gulrajani2017improved}. Specifically, WGAN-GP fulfills the 1-Lipschitz continuity of the discriminator by $\mathop{\min}\mathbb{E}_{\hat{x}\sim\pi}(||\nabla D_{\theta}(\hat{x})||_2-{\rm 1})^2$, which limits the gradient of the discriminator to 1. Although WGAN-GP solves the instability of GANs training to some extent, the assumption of optimal transport is a constrained linear programming problem. Overly strict restriction reduces the exploratory of the discriminator. In contrast, the optimal transport with the regular term mentioned is an unconstrained optimization problem. Like optimal transport corresponds to 1-Lipschitz continuity, the optimal transport with the regular term corresponds to k-Lipschitz continuity (${\rm k}\leq1$) of the discriminator, named WGAN-LP \cite{petzka2017regularization}, which is implemented by $\mathop{\min}\mathbb{E}_{\hat{x}\sim\pi}\left[\left(\mathop{\max}\{0,||\nabla D_{\theta}(\hat{x})||_2-1\}\right)^2\right]$. WGAN-LP achieves better performance by using a weaker regularization term which enforces the Lipschitz constraint of the discriminator. 
	
	WGAN-GP and WGAN-LP introduce Wasserstein distance into GANs framework. Due to the gap between limited input samples and the strict Lipschitz constraint on the whole input sample domain, the approximation of the Wasserstein distance is a challenging task. To this end, WGAN-div \cite{wu2018wasserstein} introduces a Wasserstein divergence into GANs training. The objective of WGAN-div can be smoothly derived as:
	\begin{equation}
		\mathbb{E}_{y\sim q(y)}[\varphi(y)]-\mathbb{E}_{x\sim p(x)}[\varphi(x)]+k \mathbb{E}_{\hat{x}\sim\pi}\left[ ||\varphi(\hat{x})||^p\right].
	\end{equation}
	The objective of WGAN-div is similar to WGAN-GP and WGAN-LP. It can be considered as achieving 0-Lipschitz continuity of discriminator by adopting $\mathop{\min}\mathbb{E}_{\hat{x}\sim\pi}\left[||\nabla D_{\theta}(\hat{x})||^p\right]$. 
	
	Generally, Wasserstein distance and Wasserstein divergence are reliable ways of measuring the difference between fake and real data distribution, which leads to the stable training of WGAN-based algorithms. However, a recent study \cite{stanczuk2021wasserstein} shows that the c-transform method \cite{mallasto2019well} achieves better estimation of Wasserstein divergence but leads to worse performance compared to the gradient penalty method. The results demonstrate that the success of  WGAN-based methodologies cannot truly be attributed to approximate the Wasserstein distance and the gradient penalty methods improve the performance indeed. Furthermore, some studies \cite{fedus2017many,kodali2017convergence,qi2020loss} also demonstrate that gradient penalty methods of discriminator, such as 1-GP, k-GP (${\rm k}\leq 1$), and 0-GP stabilize the training and improve the performance of GANs remarkably regardless of the loss functions. Based on these observations, stabilizing GANs training using gradient penalty is widely applied in the research community for various losses of GANs. In the rest of this section, we discuss gradient penalty methods regardless of the loss function by dividing them into three parts: \textit{1-GP}:  $\mathop{\min}\mathbb{E}_{\hat{x}\sim\pi}(||\nabla D_{\theta}(\hat{x})||-{\rm 1})^p$, \textit{k-GP (${\rm k}\leq1$)}: $\mathop{\min}\mathbb{E}_{\hat{x}\sim\pi}\left[\left(\mathop{\max}\{0,||\nabla D_{\theta}(\hat{x})||-1\}\right)^p\right]$, and \textit{0-GP}: $\mathop{\min}\mathbb{E}_{\hat{x}\sim\pi}\left[||\nabla D_{\theta}(\hat{x})||^p\right]$, where $\pi$ is the distribution of different image space (entire image space or part of image space) and $||\cdot||$ represents the norm of the gradient. Generally, the loss function of the discriminator with GP can be formulated as:
	\begin{equation}\label{GP}
		\mathcal{L}_{D}=f(\phi,\theta)+\lambda\mathcal{L}_{GP},
	\end{equation}
	where $f(\phi,\theta)$ is the uniform loss function defined in Eq (\ref{EQ:eqn1}) and $\mathcal{L}_{GP}$ is the gradient penalty regularization.
	
	\subsubsection{1-GP}
	Gulrajani et al. \cite{gulrajani2017improved} used 1-GP in WGAN-GP to train GANs. WGAN-GP uses the 2-norm gradient penalty across the entire image domain, which can be formulated as: 
	\begin{equation}\label{wgan-gp}
		\mathcal{L}_{GP}=\mathbb{E}_{\hat{x}\sim\pi}(||\nabla D_{\theta}(\hat{x})||_2-1)^2,
	\end{equation}
	where $\pi$ is the distribution of entire image space approximated by the interpolation of real distribution ($p_r$) and generated distribution ($p_g$): $\pi=t\cdot p_r+(1-t)\cdot p_g$ for $t\sim U[0,1]$. Although, WGAN-GP stabilizes the training of GANs to a great extent, the overly strict gradient penalty limits the exploratory of discriminator. To loosen the penalty, many efforts of $\pi$, $||\cdot||$, and gradient direction are proposed. 
	
	To relax the image distribution, Kodali et al. \cite{kodali2017convergence} track the training process of GANs and find that the decrease of the Inception Score (IS) is accompanied by a sudden change of the discriminator’s gradient around the real images. Authors propose DRAGAN by restricting the Lipschitz constant around the real images $\pi=p_r+\epsilon$, where $\epsilon\sim N_d(0,cI)$.
	
	In order to relax the gradient direction, Zhou et al. \cite{zhou2019towards} argue that restricting the global Lipschitz constant is unnecessary. Therefore, only maximum gradient is necessary to be penalized:
	\begin{equation}
		\mathcal{L}_{GP}=\left(\mathop{\max}\limits_{\hat{x}\sim\pi}||\nabla D_{\theta}(\hat{x})||_2-1\right)^2,
	\end{equation}
	where $\pi=t\cdot p_r+(1-t)\cdot p_g$; Furthermore, inspired by  Virtual Adversarial Training (VAT) \cite{miyato2018virtual}, D{\'a}vid et al. \cite{terjek2019virtual} propose a method, called Adversarial Lipschitz Regularization (ALR), which restricts the 1-Lipschitz continuity at $\pi=p_r\cup p_g$ in the direction of adversarial perturbation. {Adversarial perturbation direction is the most unstable direction, Restricting the 1-Lipschitz continuity to the adversarial direction means restricting only the largest Lipschitz constant, which is simpler and more efficient than the previous method.} The proposed ALP shows the SOTA performance in terms of Inception Score and Fréchet Inception Distance among non-progressive growing methods trained on CIFAR-10 dataset.
	
	Contrary to the methods which penalize the gradient in Euclidean space, Adler et al. \cite{adler2018banach} extended the $L_p(p=2)$ space with gradient penalty to Banach space that contains the $L_p$ space and Sobolev space. For the Banach space B, the Banach norm $||.||_{B^*}$ is defined as:  
	\begin{equation}
		||x^*||_{B^*}=\sup\limits_{x\in B}\frac{x^*(x)}{||x||_B}.
	\end{equation}
	Thus, the gradient penalty of Banach wasserstein GAN can be expressed as:
	\begin{equation}
		\mathcal{L}_{GP}=\mathbb{E}_{\hat{x}\sim\pi}(||\nabla D_{\theta}(\hat{x})||_{B^*}-1)^2,
	\end{equation}
	where $\pi=t\cdot p_r+(1-t)\cdot p_g$. {Banach wasserstein GAN expands the Lipschitz continuity into Banach space containing both $L_p$ space and Sobolev space, which has more restriction than wasserstein GAN.}
	\subsubsection{k-GP (${\rm k}\leq1$)}
	{k-GP (${\rm k}\leq1$) was first tested by Gulrajani et al. \cite{gulrajani2017improved} and named one sided gradient penalty. It uses the 2-norm gradient penalty across the entire image domain, which is formulated as: }
	\begin{equation}\label{wgan-lp}
		\mathcal{L}_{GP}=\mathbb{E}_{\hat{x}\sim\pi}\left[\left(\mathop{\max}\{0,||\nabla D_{\theta}(\hat{x})||_2-1\}\right)^2\right],
	\end{equation}
	where $\pi$ is the distribution of entire image space approximated by the interpolation of real distribution ($p_r$) and generated distribution ($p_g$): $\pi=t\cdot p_r+(1-t)\cdot p_g$ for $t\sim U[0,1]$. Inspired by the optimal transport with the regular term, Petzka et al. \cite{petzka2017regularization} also used k-GP (${\rm k}\leq1$) to training GANs named WGAN-LP. Furthermore, Xu et al \cite{xu2021generalized} show a more general dual form of the Wasserstein distance compared to KR duality (mentioned in section 2.4), named Sobolev duality, which relaxes the Lipschitz constraint but still maintains the favorable gradient
	property of the Wasserstein distance. Authors also show that the KR duality is a special case of the proposed Sobolev duality. Based on the Sobolev duality, the relaxed gradient penalty of the proposed SWGAN is formulated as:
	
	\begin{equation}\label{wgan-sp}
		\mathcal{L}_{GP}=\mathbb{E}_{\hat{x}\sim\pi}\left[\left(\mathop{\max}\{0,||\nabla D_{\theta}(\hat{x})||^2-1\}\right)^2\right],
	\end{equation}
	where $\pi=t\cdot p_r+(1-t)\cdot p_g$ for $t\sim U[0,1]$. It is clear that  above three method have the same form of gradient penalty. Interestingly, different relaxation methods yield the same form of regularization.
	\subsubsection{0-GP}
	\begin{table}
		\caption{The Gradient penalty of the Discriminator. $\mu$ and $v$ are real and generated distribution, respectively.}
		\footnotesize
		\label{table:gradient penalty}
		\centering
		\begin{tabular}{c | c | c | c }	
			\toprule
			\midrule
			{ Method}& { $\mathcal{L}_{GP}$}& $\pi$&Lipschitz continuity	 \\
			\hline
			WGAN-GP \cite{gulrajani2017improved}&$\mathbb{E}_{\hat{x}\sim\pi}(||\nabla D_{\theta}(\hat{x})||_2-1)^2$&$t\cdot p_r+(1-t)\cdot p_g$&$||D_{\theta}||_{Lip}\to1$\\
			\hline
			DRAGAN \cite{kodali2017convergence}&$\mathbb{E}_{\hat{x}\sim\pi}(||\nabla D_{\theta}(\hat{x})||_2-1)^2$&$p_r+\epsilon$&$||D_{\theta}||_{Lip}\to1$\\
			\hline 
			Max-GP \cite{zhou2019towards}&$\left(\mathop{\max}\limits_{\hat{x}\sim\pi}||\nabla D_{\theta}(\hat{x})||_2-1\right)^2$&$t\cdot p_r+(1-t)\cdot p_g$&$||D_{\theta}||_{Lip}\to1$\\
			\hline
			ALP \cite{terjek2019virtual}&$\mathbb{E}_{\hat{x}\sim\pi}(||\nabla D_{\theta}(\hat{x})||_2-1)^2$&$p_r\cup p_g$&$||D_{\theta}||_{ALP-Lip}\to1$\\
			\hline
			Banach-GP \cite{adler2018banach}&$\mathbb{E}_{\hat{x}\sim\pi}(||\nabla D_{\theta}(\hat{x})||_{B^*}-1)^2$&$t\cdot p_r+(1-t)\cdot p_g$&$||D_{\theta}||_{Lip}\to1$\\
			\hline
			WGAN-LP \cite{petzka2017regularization}&$\mathbb{E}_{\hat{x}\sim\pi}\left[\left(\mathop{\max}\{0,||\nabla D_{\theta}(\hat{x})||_2-1\}\right)^2\right]$&$t\cdot p_r+(1-t)\cdot p_g$&$||D_{\theta}||_{Lip}\leq1$\\
			\hline
			SWGAN \cite{xu2021generalized}&$\mathbb{E}_{\hat{x}\sim\pi}\left[\left(\mathop{\max}\{0,||\nabla D_{\theta}(\hat{x})||^2-1\}\right)^2\right]$&$t\cdot p_r+(1-t)\cdot p_g$&$||D_{\theta}||_{Lip}\leq1$\\
			\hline
			zc-GP \cite{wu2018wasserstein,mescheder2018training,zhang2018wasserstein,9851855}&$\mathbb{E}_{\hat{x}\sim\pi}||\nabla D_{\theta}(\hat{x})||^2_2$&$p_r\cup p_g$&$||D_{\theta}||_{Lip}\to0$\\
			\hline
			GAN-QP \cite{su2018gan}&$\mathcal{L}_{GP}=\mathbb{E}_{x_r,x_g\sim\pi}\frac{\left(D_{\theta}(x_r)-D_{\theta}(x_f)\right)^2}{||x_r-x_f||}$&$\pi=p_r\cdot p_g$&$\frac{\left(D_{\theta}(x_r)-D_{\theta}(x_f)\right)^2}{||x_r-x_f||}\to 0$\\
			\hline
			ZP-Max \cite{zhou2019lipschitz}&$\mathop{\max}\limits_{\hat{x}\sim\pi}||\nabla D_{\theta}(\hat{x})||^2_2$&$t\cdot p_r+(1-t)\cdot p_g$&$||D_{\theta}||_{Lip}\to0$\\
			\hline
			ZP \cite{thanh2019improving}&$\mathbb{E}_{\hat{x}\sim\pi}||\nabla D_{\theta}(\hat{x})||^2_2$&$t\cdot p_r+(1-t)\cdot p_g$&$||D_{\theta}||_{Lip}\to0$\\
			\bottomrule
		\end{tabular}
	\end{table} 
	
	Wu et al. \cite{wu2018wasserstein} used 0-GP, and proposed Wasserstein divergence. According to \cite{evans1997partial}, Wasserstein divergence is solved by minimizing:
	\begin{equation}
		\label{0-gp}
		\mathcal{L}_{GP}=\mathbb{E}_{\hat{x}\sim\pi}||\nabla D_{\theta}(\hat{x})||^2,
	\end{equation}
	where $\pi$ is both the real distribution ($p_r$) and the generated distribution ($p_g$): $\pi=p_r\cup p_g$. Furthermore, Mescheder et al. \cite{mescheder2018training} also demonstrate that the optimization of unregularized GAN is not always locally convergent and some simplified zero centered gradient penalty (zc-GP) techniques, implemented by minimizing Eq (\ref{0-gp}), can be used to achieve local convergence of GANs. Li et al. \cite{9851855} introduce the adversarial training to discriminator training, which is turned out to be an adaptive 0-GP. 
	
	Besides, some other 0-GP methods \cite{su2018gan,zhang2018wasserstein,zhou2019lipschitz,thanh2019improving,9851855} are derived by different theoretical derivations. For instance, Su et al. \cite{su2018gan} propose a Quadratic Potential (QP) for GANs training with the following formulation:
	\begin{equation}
		\mathcal{L}_{GP}=\mathbb{E}_{x_r,x_g\sim\pi}\frac{\left(D_{\theta}(x_r)-D_{\theta}(x_f)\right)^2}{||x_r-x_f||},
	\end{equation}
	where $\pi$ is the joint distribution of the real and generated distributions: $\pi=p_r\cdot p_g$; Zhang et al. \cite{zhang2018wasserstein} combine a Total Variational (TV) regularizing term into the training of GANs, that is $|D_{\theta}(x_r)-D_{\theta}(x_f)-\delta|$. According to \cite{zhang2018wasserstein}, the TV term can be approximated by Eq (\ref{0-gp}), which is exhilarating; Zhou et al. \cite{zhou2019lipschitz} propose the Lipschitz GANs, with the maximum of the gradients penalty for guaranteeing the gradient informativeness:
	\begin{equation}
		\mathcal{L}_{GP}=\mathop{\max}\limits_{\hat{x}\sim\pi}||\nabla f(\hat{x})||^2_2,
	\end{equation}
	where $\pi=t\cdot p_r+(1-t)\cdot p_g$; Thanh-Tung et al. \cite{thanh2019improving} also propose the 0-GP with gradients penalty at $\pi=t\cdot p_r+(1-t)\cdot p_g$:
	\begin{equation}
		\mathcal{L}_{GP}=\mathbb{E}_{\hat{x}\sim\pi}||\nabla f(\hat{x})||^2_2.
	\end{equation}
	
	In summary, gradient penalty techniques are widely used in the GANs training to achieve Lipschitz continuity of discriminator. As shown in Table \ref{table:gradient penalty}, many techniques are proposed based on different theories and phenomena. But to the best of our knowledge, there is no fair and comprehensive work comparing the performance of these gradient penalty methods.
	To compare the performance of various methods intuitively, a comparative experiment on CIFAR-10 and CIFAR-100 datasets is conducted\footnote{The base framework comes from wgan-gp in \url{https://github.com/kwotsin/mimicry}}. The results of FID \cite{heusel2017gans} for various gradient penalty methods with different loss functions are shown in Table \ref{table: experiments results}. The results validate the conclusion in studies \cite{stanczuk2021wasserstein,fedus2017many,kodali2017convergence}, that the Lipschitz constraint of discriminator may improve the performance and stability of GANs training regardless of the statistical distance used as a loss function. All gradient penalty methods improve the performance of GANs upon all three loss functions. Among them, zc-GP \cite{wu2018wasserstein,mescheder2018training,zhang2018wasserstein} obtains the best performance and is widely used in SOTA methods as illustrated in Table \ref{table:SOTA }.

	\begin{table*}
		\tiny
		\caption{ FID results on the CIFAR-10 and CIFAR-100 datasets for various gradient penalty methods with different GAN losses.}
		\centering
		\label{table: experiments results}
		\begin{tabular}{cccccccccc}
			\toprule
			\multirow{2}{*}{Dataset}&\multirow{2}{*}{Loss} & \multicolumn{8}{c}{\makecell*[c]{Gradient Penalty Methods}} \\ 
			\cline{3-10} 
			&& \makecell*[c]{None}&GP\cite{gulrajani2017improved}&DRAGAN \cite{kodali2017convergence}&MAX-GP \cite{zhou2019towards}&LP\cite{petzka2017regularization}&zc-GP \cite{wu2018wasserstein,mescheder2018training,zhang2018wasserstein}&ZP-MAX \cite{zhou2019lipschitz}&ZP \cite{thanh2019improving} \\ 
			\midrule
			\multirow{3}{*}{CIFAR-10}&
			GAN\cite{goodfellow2014generative}& 42.41& 23.45&20.98& 26.65&22.9&\textbf{19.39}&24.38&23.96 \\
			&WGAN\cite{arjovsky2017wasserstein}&290&30.38& 29.53&37.21&28.31&\textbf{26.99}&31.28&30.19\\
			&Hinge\cite{lim2017geometric}&58.34&21.19&21.77&25.4&20.79&\textbf{18.75}&23.1&22.58\\
			\cline{1-10} 
			\specialrule{0em}{2pt}{2pt}
			\multirow{3}{*}{CIFAR-100}&
			GAN\cite{goodfellow2014generative}&44.5&25.76& 25.37&24.29&23.82&\textbf{21.81}&26.27&25.38 \\
			&WGAN\cite{arjovsky2017wasserstein}&244&32.28&31.93&38.71&32.19&\textbf{29.12}&39.75&37.8 \\
			&Hinge\cite{lim2017geometric}&59.43&25.13&25.42&28.34&23.67&\textbf{21.55}&26.19&26.06\\
			\bottomrule
		\end{tabular}
	\end{table*}
	\subsection{Weight Normalization and Weight Regularization}
	WGAN is a popular and important generative adversarial network. From the optimal transport introduced on the Section \ref{sect:A-1} of the Supplementary Online-only Material, to obtain $W_c(\mu,\upsilon)=\mathop{\max}\limits_{||\varphi_\xi||_L\leq 1}\left\{\int_X \varphi_\xi \mathrm{d}\mu\ -\int_Y \varphi_\xi \mathrm{d}\upsilon\right\}$, the discriminator must satisfy the 1-Lipschitz continuity. According to the Section \ref{sect:A-3} in the Supplementary Online-only Material, the spectral norm $||W||_2$ can be used to represent the Lipschitz constant $\mathrm{C'}$. The Lipschitz continuity is achieved by normalizing the spectral norm of the weight, approximately. Hence, \textit{Weight Normalization} and \textit{Weight Regularization} can also be used to enable the Lipschitz continuity of the discriminator.
	
	\subsubsection{Weight Normalization}
	Spectral norm of the weight and the Lipschitz constant express the same concept. Therefore, weight normalization is another method to achieve Lipschitz continuity. {More important, weight normalization methods are Non-sampling-based, which don’t have the lack of support problem in contrast to gradient penalties.} Spectral normalization of the weight limits the Lipschitz constant to 1. Certainly, upper bound of the spectral norm can be used to normalize the weights, achieving $k (k\leq1)$ Lipschitz continuity. The following lemmas put forward some upper bounds of the spectral norm.
	\newenvironment{lemma3}{{\indent\it \textbf{Lemma 3.1:}}}{\hfill \par}
	
	\begin{lemma3}
		\textit{
			If $\lambda_1\leq\lambda_2\leq\cdots\leq\lambda_M$
			are the eigenvalues of the $W^\top W$, then the spectral norm$||W||_2=\sqrt{\lambda_M}$; The Frobenius norm$||W||_F=\sqrt{\sum_{i=1}^{M}\lambda_i}$ 
		}
	\end{lemma3}
	\newenvironment{lemma4}{{\indent\it \textbf{Lemma 3.2:}}}{\hfill \par}
	\newenvironment{lemma5}{{\indent\it \textbf{Lemma 3.3:}}}{\hfill \par}
	\newenvironment{proof3}{{\indent\it Proof 3.1:}}{\hfill $\square$\par}
	\newenvironment{proof4}{{\indent\it Proof 3.2:}}{\hfill $\square$\par}
	\newenvironment{proof5}{{\indent\it Proof 3.3:}}{\hfill $\square$\par}
	\begin{proof3}
		See \cite{mathias1990spectral} and \cite{miyato2018spectral}
	\end{proof3}
	\begin{lemma4}
		\textit{
			For a $n\times m$ matrix, $||W||_1=\mathop{\max}\limits_{j}\sum_{i=1}^{n}|a_{i,j}|$, $ ||W||_\infty=\mathop{\max}\limits_{i}\sum_{j=1}^{m}|a_{i,j}|$, then $||W||_2\leq\sqrt{||W||_1||W||_\infty}$
		}
	\end{lemma4}
	\begin{proof4}
		See \cite{mathias1990spectral}
	\end{proof4}
	\begin{lemma5}
		\textit{
			For a $n\times m$ matrix, 
			$||W||_F=\sqrt{\left(\sum_{j=1}^{m}\sum_{i=1}^{n}|a_{i,j}|^2\right)}$, then $||W||_2\leq||W||_F$
		}
	\end{lemma5}
	\begin{proof5}
		See \cite{mathias1990spectral}
	\end{proof5}
	
	1-Lipschitz continuity can be expressed by the spectral normalization. Miyato et al. \cite{miyato2018spectral} control the Lipschitz constant through spectral normalization $W_\sigma=\frac{W}{||W||_2}$ of each layer for D, leading to a better result than WGAN-GP. {Practically, the power iteration method is used as a fast approximation for the spectral norm ($||W||_2$).} Similarly, according to the optimal transport with regular term, Lipschitz constant of discriminator should be less than or equal to 1. Correspondingly, upper bound of the spectral norm can be utilized to normalize the weight ($||W_\sigma||_2\leq1$), achieving $k (k\leq1)$ Lipschitz continuity. In terms of Lemma 3.2 and Lemma 3.3, $\sqrt{||W||_1||W||_\infty}$ and Frobenius norm ($||W||_F$) are simple upper bound of the spectral norm ($||W||_2$) and can be used to normalize the weight. For example, Zhang et al. \cite{zhang2019spectral} use the $\sqrt{||W||_1||W||_\infty}$, seeking for an approximation of the spectral norm that is easy to calculate. Miyato et al. \cite{miyato2018spectral} explain that the Frobenius norm is a restriction on all eigenvalues. It is different from the spectral norm, which only constrains the maximum eigenvalue. Authors conjecture that Frobenius normalization affects the network's ability to express, but no experiments are reported to compare it with the spectral normalization. Liu et al. \cite{liu2019spectral} find that the mode collapse is often accompanied by the collapse of the eigenvalue of the discriminator. Because the spectral normalization only limits the maximum eigenvalue, and the eigenvalue collapse means the remaining eigenvalues suddenly decrease. Therefore, authors adopt the following methods to prevent the collapse of the eigenvalues:
	\begin{equation}
		W_{\sigma}=\frac{W+\nabla W}{||W||_2}=\frac{W}{||W||_2}+\frac{\nabla W}{||W||_2}.
	\end{equation}
	The results demonstrate that this method effectively prevents mode collapses. Although the experiments are reported in this study, but it misses theoratical proofs. Therefore the relationship between the matrix eigenvalues and GAN performance is not clear.
	
	Few researches focus on weight normalization as demonstrated in Table \ref{table:norm normalization}. Among these studies, spectral normalization is widely applied in some SOTA methods, as demonstrated in Section 7.
	\begin{table}
	\caption{The summary of the weight normalization and wight regularization.}
	\label{table:norm normalization}
	\centering
	\begin{tabular}{c | c | c  }	
		\toprule
		\midrule
		{ Method}&Implementation &Motivation	 \\
		\hline
		Spectral normalization (SN) \cite{miyato2018spectral}&$W_\sigma=W/||W||_2$&$||D||_{Lip}\to1$\\
		\hline
		F normalization \cite{miyato2018spectral}&$W_\sigma=W/||W||_F$&$||D||_{Lip}\leq1$\\
		\hline
		Mixed normalization \cite{zhang2019spectral}&$W_\sigma=W/\sqrt{||W||_1||W||_\infty}$&$||D||_{Lip}\leq1$\\
		\hline
		Spectral increment normalization \cite{liu2019spectral}&$W_\sigma=W/||W||_2+\nabla W/||W||_2$&$||D||_{Lip}\to1$\\
		\bottomrule
	\end{tabular}
\end{table}	
	
	\subsubsection{Weight Regularization}
	Compared with spectral normalization similar to 1-GP, spectral regularization is similar to the 0-GP. Kurach et al. \cite{kurach2018large} use the $\mathcal{L}_R=||W||_2$ to regularize the loss function. Zhou et al. \cite{zhou2018lp} also use the $L_P$-norm ($P=1,F,\infty$) to regularize the discriminator. However, these studies have worse performance than weight normalization and did not catch much attention among researchers.
	
	\subsection{{Gradient Normalization}}
	Gradient normalization is also a popular method to impose the Lipschitz constraint on the discriminator. As we all know, 1-Lipschitz constraint can be implemented by $\mathop{\min}\mathbb{E}_{{x}\sim\pi}(||\nabla_{x} D_{\theta}({x})||_2-1)^2$ to let the gradient of the discriminator ($||\nabla_{{x}} D_{\theta}({x})||_2$) equal to 1. Therefore, \cite{wu2021gradient,bhaskara2022gran} control the Lipischitz constant through gradient normalization $\hat{D}_{\theta}(x)=\frac{D_{\theta}(x)}{||\nabla_{{x}}D_{\theta}({x})||_2}$ for $D_{\theta}$. Accordingly, the gradient of $\hat{D}_{\theta}$ can be represented as $||\nabla_{{x}} \hat {D}_{\theta}({x})||_2=||\nabla_{{x}}\left(\frac{D_{\theta}(x)}{||\nabla_{{x}}D_{\theta}({x})||_2}\right)||_2$, equaling to 1. To ensure the boundedness, different studies have different implementation. For instance, \cite{wu2021gradient} adopts $\hat{D}_{\theta}(x)=\frac{D_{\theta}(x)}{||\nabla_{{x}}D_{\theta}({x})||_2+D_{\theta}(x)}$ and \cite{bhaskara2022gran} adopts $\hat{D}_{\theta}(x)=\frac{D_{\theta}(x)}{||\nabla_{{x}}D_{\theta}({x})||_2+\epsilon}$.  Extensive experiments \cite{wu2021gradient,bhaskara2022gran} demonstrate that both implementation of gradient normalization attain significant performance gains comparing to gradient penalty, weight normalization, and weight regularization.
	\subsection{{Summary}}
	{As mentioned above, weight clipping, gradient penalty, weight regularization, weight normalization, and gradient normalization all could enable Lipschitz continuity of the discriminator. However, what are the advantages and disadvantages of investigated techniques? Imposing the Lipschitz constraint on the discriminator can be characterized by three properties \cite{wu2021gradient}. 1) \textit{model-} or \textit{module-wise} \textit{constraint}. Model-wise constraint is defined as methods that constraint objective depends on full model, while module-wise constraint is defined as methods that constraint objective depends on layers. Generally, model-wise constraint is better since module-wise constraint is strict, which limits
		the layer capacities and reduces the power of discriminator. 2) \textit{sampling-based} or \textit{non-sampling} \textit{-based} \textit{constraint}. Sampling-based constraint is defined as requiring sampling data during usage, while non-sampling-based constraint depends on the model, not data sampling. Generally, non-sampling-based constraint performs better since Lipschitz constraint should be fulfilled on the entire data manifold, not only sampling data. 3) \textit{Hard} or \textit{soft constraint}. {The accurate constraint of Lipschitz continuity is defined as hard constraint and the converse to be soft constraint. Hard constraint has achieved the exact Lipschitz continuity through limiting the spectral norm, which is expected to perform better. While soft constraint only obtain the Lipschitz continuity approximatively through optimization.} Table \ref{table:lipschitz continuity} summarizes the properties of different technologies, from which gradient normalization is a model-wise, non-sampling-based, and hard constraint method.}
	
	\begin{table}
		\caption{Summary of different regularization and normalization technologies for imposing Lipschitz continuity.}
		\label{table:lipschitz continuity}
		\centering
		\begin{tabular}{c | c | c | c }	
			\toprule
			\midrule
			Method& Model-wise&Non-sampling-based&Hard\\
			\hline
			Weight Clipping&&\checkmark&\\
			\hline
			Gradient Penalty&\checkmark&&\\
			\hline
			Weight Regularization&&\checkmark&\\
			\hline
			Weight Normalization&&\checkmark&\checkmark\\
			\hline
			Gradient Normalization&\checkmark&\checkmark&\checkmark\\
			\bottomrule
		\end{tabular}
	\end{table}

	\section{Regularization and Normalization of "Training dynamics"}
	Assuming the objectives of GANs are convex-concave, some studies have proposed the global convergence of GANs \cite{nowozin2016f,gidel2018variational}. However, these theoretical convergence analyses are only applicable to the GANs with the optimal discriminator. Therefore, some studies focus on analyzing the local convergence of GANs. According to Nagarajan et al. \cite{nagarajan2017gradient} and Mescheder et al. \cite{mescheder2017numerics}, under some assumptions, GANs dynamics are locally convergent. However, if these assumptions are not satisfied, especially if the data distributions are not continuous, GANs dynamics do not always converge locally unless some regularization techniques are used.
	
	We review {Jacobian regularization} techniques \cite{mescheder2017numerics,nagarajan2017gradient} in this section, which minimize the Jacobian matrix to achieve local convergence. With the same motivation, Mescheder et al. \cite{mescheder2018training} propose a simplified gradient penalties method, named zero-centered gradient penalties (zc-GP), that guarantees the local convergence under suitable assumptions. Since it is similar to 0-GP, we cover it in Section 4.
	\subsection{Jacobian Regularization}
	In \textit{Proposition 2.2} of Section 2: absolute values of all eigenvalues of the Jacobian matrix ($v^{'}(\phi,\theta)$) are expected to be less than 1 at the fixed point, which is equivalent to the real part of the eigenvalue being negative. Additionally, the learning rate must be relatively low \cite{mescheder2017numerics}. To meet these requirements, Mescheder et al. \cite{mescheder2017numerics} used the Consensus Optimization (ConOpt) to make the real part of the eigenvalue negative. Its regularized updates are:
	\begin{equation} \label{eqn1}
		\begin{split}
			\phi^{(k+1)}=\phi^{(k)}+h\nabla_\phi\left(- f(\phi^{(k)},\theta ^{(k)})-\gamma L(\phi^k,\theta^k)\right),\\
			\theta^{(k+1)}=\theta^{(k)}+h\nabla_\theta\left(f(\phi^{(k)},\theta ^{(k)})-\gamma L(\phi^k,\theta^k)\right),
		\end{split}
	\end{equation}
	where $L(\phi^k,\theta^k)=\frac{1}{2}||v(\phi^k,\theta^k)||^2=\frac{1}{2}\left(||\nabla_\phi f(\phi^k,\theta^k)||^2+||\nabla_\theta f(\phi^k,\theta^k)||^2\right)$ is the regularization of the Jacobian matrix.
	
	Apart from \cite{mescheder2017numerics}, Nagaraja et al. \cite{nagarajan2017gradient} also analyze the relationship between local convergence of GANs and all eigenvalues of the Jacobian
	of the gradient vector field. Authors prove the local convergence for absolutely continuous generator and data distributions under certain regularity assumptions. This requires the loss function of the GANs to be strictly concave, which is not the case for some GANs. Based on this, a simple regularization technology that regularized the generator using the gradient of the discriminator is proposed by Nagaraja et al. \cite{nagarajan2017gradient}. The regularized updates for the generator can be expressed as:
	\begin{equation} \label{eqn1}
		\phi^{(k+1)}=\phi^{(k)}-h\nabla_\phi f(\phi^{(k)},\theta ^{(k)})-\frac{1}{2}h\gamma\nabla_\phi||\nabla_\theta f(\phi^{k},\theta^{k})||^2.
	\end{equation}
	Herein, the update of the discriminator is similar to SimGD. Furthermore, Nie et al. \cite{nie2019towards} propose a method that only regularizes the discriminator. The regularized update of the discriminator in this case is given by:
	\begin{equation} \label{eqn1}
		\theta^{(k+1)}=\theta^{(k)}+h\nabla_\theta f(\phi^{(k)},\theta ^{(k)})-\frac{1}{2}h\gamma\nabla_\theta||\nabla_\phi f(\phi^{k},\theta^{k})||^2.
	\end{equation}
	The update of the generator is the same as SimGD. Nie et al. \cite{nie2019towards} propose JAcobian REgularization (JARE) that regularizes both the generator and the discriminator. The regularized updates for the generator and the  discriminator are:
	\begin{equation} \label{eqn1}
		\begin{split}
			\phi^{(k+1)}=\phi^{(k)}-h\nabla_\phi f(\phi^{(k)},\theta ^{(k)})-\frac{1}{2}h\gamma\nabla_\phi||\nabla_\theta f(\phi^{k},\theta^{k})||^2,\\
			\theta^{(k+1)}=\theta^{(k)}+h\nabla_\theta f(\phi^{(k)},\theta ^{(k)})-\frac{1}{2}h\gamma\nabla_\theta||\nabla_\phi f(\phi^{k},\theta^{k})||^2.
		\end{split}
	\end{equation}
	The key difference between JARE and ConOpt is that JARE does not contain the Hessians $\nabla^2_{\phi,\phi}f(\phi^{k},\theta^k)$ and $\nabla^2_{\theta,\theta}f(\phi^{k},\theta^k)$ in the regularization term. 
	
	{There are several Jacobian regularization methods that have been proposed to deal with the training instabilities of GANs. What is the difference of them? Nie et al. \cite{nie2019towards} consider a simple toy example to analyse the convergence of GANs. There may exist two factors of the Jacobian in the GANs dynamics simultaneously that destroy the GANs training: (i) the Phase Factor, i.e., the Jacobian has complex eigenvalues with a large imaginary-to-real ratio; (ii) the Conditioning Factor, i.e., the Jacobian is ill-conditioned. According to the toy example,
		Only Regularizing Generator \cite{nagarajan2017gradient}, Only Regularizing Discriminator \cite{nie2019towards}, and ConOpt \cite{mescheder2017numerics} could only alleviate the impact of the Phase Factor but not alleviating the impact of the Conditioning Factor. However, JARE \cite{nie2019towards} can address both factors by construction}.\footnote{Intuitively, a reason for not introducing Hessians in JARE \cite{nie2019towards} is to avoid the risk of reversing the gradient flows, which may diverge the GAN training dynamics (see Appendix C in \cite{nie2019towards} for a detailed explanation).}
	
	The above discussions of local convergence during GANs training involve a premise: absolutely continuous data and generator distributions. Indeed, the assumption of absolute continuity is not true for common cases of GANs, where both distributions, specially the data distribution, may lie on lower-dimensional manifolds \cite{arjovsky2017principled}. More generally, Mescheder et al. \cite{mescheder2018training} extend the convergence proof by \cite{nagarajan2017gradient} to the case where the generator and data distribution do not locally have the same support. Based on this, a simplified zero-centered gradient penalties (zc-GP) method is proposed, which guarantees the local convergence under suitable assumptions. Zc-GP is obtained from the training dynamics, which is similar to 0-GP methods mentioned in Section 4.
	
	{Furthermore, there are also other literature studies \cite{qin2020training,wang2019solving,schafer2019competitive} analyze the training of GANs through other tools. For instance, \cite{schafer2019competitive} introduces a novel algorithm, competitive gradient descent (CGD), that is a natural extension of gradient descent to the competitive setting. Different from gradient descent ascent (GDA) in Eq (\ref{eq:eqn27}), CGD does not need to reduce the stepsize to match the increase of the interactions to avoid divergence. Specifically, CGD introduces an equilibrium term that lets each player prefer strategies that are less vulnerable to the actions of the other player. \cite{wang2019solving} also elucidates the cause of undesirable convergence of GDA is leader's (discriminator) gradient step takes the system away from the ridge, which has undesirable convergence properties and requires using very small learning rates to converge. To mitigate this, Follow-the-Ridge (FR) term ($\mathbf{H}_{\mathbf{\theta} \mathbf{\theta}}^{-1} \mathbf{H}_{\mathbf{\theta} \mathbf{\phi}} \nabla_{\mathbf{\phi}} f\left(\mathbf{\phi}^{(k)}, \mathbf{\theta}^{(k)}\right)$) has been added to the updating of the discriminator. \cite{qin2020training} studies the continuous-time dynamics induced by GANs training. In this perspective, instabilities in training GANs arise from the integration error in discretizing the continuous dynamics. It treats GANs training as solving ODEs and shows that higher-order solvers lead to better convergence.}
	
	\subsection{Summary}
	In summary, Jacobian regularization techniques are obtained from the training dynamics of GANs, which are used for achieving local convergence and stabilizing training. The summary of the Jacobian regularization methods is demonstrated in Table \ref{table:Jacobian regularization}. Jacobian regularization is similar to the Gradient penalty in terms of update form. In general, zc-GP is used in many SOTA methods, as demonstrated in Section 7.
	\begin{table}
		\caption{The summary of the Jacobian regularization.}
		\tiny
		\label{table:Jacobian regularization}
		\centering
		\begin{tabular}{c|c|c}	
			\toprule
			\midrule
			Method& regularized updates of generator ($\phi^{(k+1)}$)&regularized updates of discriminator ($\theta^{(k+1)}$)	 \\
			\hline
			SimGD \cite{goodfellow2014generative}&$\phi^{(k)}-h\nabla_\phi f(\phi^{(k)},\theta ^{(k)})$&$\theta^{(k)}+h\nabla_\theta f(\phi^{(k)},\theta ^{(k)})$\\
			\hline
			ConOpt \cite{mescheder2017numerics}&$\phi^{(k)}-h\nabla_\phi f(\phi^{(k)},\theta ^{(k)})-\frac{1}{2}h\gamma\nabla_\phi||v(\phi^{(k)},\theta^{(k)})||^2$&$\theta^{(k)}+h\nabla_\theta f(\phi^{(k)},\theta ^{(k)})-\frac{1}{2}h\gamma\nabla_\theta||v(\phi^{(k)},\theta^{(k)})||^2$\\
			\hline
			Generator \cite{nagarajan2017gradient}&$\phi^{(k)}-h\nabla_\phi f(\phi^{(k)},\theta ^{(k)})-\frac{1}{2}h\gamma\nabla_\phi||\nabla_\theta f(\phi^{(k)},\theta^{(k)})||^2$&$\theta^{(k)}+h\nabla_\theta f(\phi^{(k)},\theta ^{(k)})$\\
			\hline
			Discriminator \cite{nie2019towards}&$\phi^{(k)}-h\nabla_\phi f(\phi^{(k)},\theta ^{(k)})$&$\theta^{(k)}+h\nabla_\theta f(\phi^{(k)},\theta ^{(k)})-\frac{1}{2}h\gamma\nabla_\theta||\nabla_\phi f(\phi^{(k)},\theta^{(k)})||^2 $\\
			\hline
			JARE \cite{nie2019towards}&$\phi^{(k)}-h\nabla_\phi f(\phi^{(k)},\theta ^{(k)})-\frac{1}{2}h\gamma\nabla_\phi||\nabla_\theta f(\phi^{(k)},\theta^{(k)})||^2$&$\theta^{(k)}+h\nabla_\theta f(\phi^{(k)},\theta ^{(k)})-\frac{1}{2}h\gamma\nabla_\theta||\nabla_\phi f(\phi^{(k)},\theta^{(k)})||^2 $\\
			\hline
			zc-GP \cite{mescheder2018training}&$\phi^{(k)}-h\nabla_\phi f(\phi^{(k)},\theta ^{(k)})$&$\theta^{(k)}+h\nabla_\theta f(\phi^{(k)},\theta ^{(k)})-\frac{1}{2}h\gamma\nabla_\theta||\nabla D_{\theta}(x)||^2 $\\
			\bottomrule
		\end{tabular}
	\end{table}

	\section{Regularization and Normalization of "Other methods"}
	In addition to the three groups mentioned above, this section discusses and summarizes the remaining regularization and normalization techniques, namely, \textit{Layer Normalization} and \textit{Inverse Gradient Penalty}. \textit{Layer Normalization}  consists of unconditional-based layer normalization and conditional-based layer normalization, the former inspired by supervised learning is used to accelerate training, but its impact on the GANs is small and sometimes 
	drops the performance, while the latter is used in the conditional generation and significantly improves the performance of conditional generation; On the other hand, \textit{Inverse Gradient Penalty} mitigates mode collapse by maximizing the Lipschitz constant of the generator.
	\subsection{Layer Normalization}
	Data in machine learning is expected to be independent and identically distributed ($i.i.d$). However, in terms of deep learning, because of the Internal Covariate Shift (ICS) \cite{ioffe2015batch}, inputs of each neuron do not satisfy the $i.i.d$, making the training of the deep neural networks hard and unstable. Layer normalization\footnote{\label{ft:10} layer normalization is different from the Layer Normalization (LN), where layer normalization is a general term for a class of methods such as BN, LN. } has been proposed to avoid such problems. The general form of the layer normalization is (the difference between the normalization methods lies in the choice of $h$ and the calculation of $\mathbb{E}[h]$ and $var[h]$):
	\begin{equation}
		h_N=\frac{x-\mathbb{E}[h]}{\sqrt{var[h]+\epsilon}}\cdot\gamma+\beta.
	\end{equation}
	
	For GANs, the layer normalization is divided into two parts: {\textit{unconditional-based layer normalization}} and {\textit{conditional-based layer normalization}}. Unconditional-based layer normalizations are used for unconditional generation similar to the other deep neural networks. On the other hand, conditional-based layer normalizations are used for the generator of the conditional generation, where the shift and scale parameters ($\gamma, \beta$) depend on the condition information, as given below:
	\begin{equation}
		h_N=\frac{x-\mathbb{E}[h]}{\sqrt{var[h]+\epsilon}}\cdot\gamma(c)+\beta(c).
	\end{equation}
	
	\subsubsection{Unconditional-based layer Normalization}
	Unconditional-based layer normalization is used for both the generator and discriminator with the same motivation as in other deep neural networks. Ioffe et al. \cite{ioffe2015batch} proposed the first normalization for neural networks, namely, Batch Normalization (BN). Batch normalization adopts the data of the mini-batch to compute the mean and variance, making the data distribution of each mini-batch approximately the same. Miyato et al. \cite{miyato2018spectral} used the BN in GANs. BN normalizes at the mini-batch level, which destroys the difference between pixels during the generation on account of image generation being a pixel-level task. {Therefore, Batch Norm can be less applicable to style transfer and can’t be used with gradient penalty
		methods since the gradient would be dependent on multiple inputs.} Contrary to BN which normalizes the same channel with different images, Layer Normalization\textsuperscript{\ref {ft:10}} (LN) \cite{ba2016layer} normalizes different channels of a single image that also destroys the diversity between channels for the pixel-by-pixel generative model \cite{miyato2018spectral}. Instance Normalization (IN) \cite{ulyanov2016instance} has also been proposed for style transformation that is adopted for a single channel of a single image. Moreover, Group Normalization (GN) \cite{wu2018group} sits between LN and IN, which first divides the channel into many groups, and normalizes different groups of a single image. Compared to normalization of input of neural networks in BN, LN, IN and GN, Weight Normalization (WN) \cite{salimans2016weight} normalizes the weight matrix of neural networks. Miyato et al. \cite{miyato2018spectral} also used this normalization in GANs. 
	
	In summary, unconditional-based layer normalization in GANs is similar to other neural networks. The related summaries are shown in Table \ref{table:layer normalization}. To the best of our knowledge, no study compares the performance of these methods, therefore, we demonstrate the FID results\footnote{The base framework comes from the SNGAN in \url{https://github.com/kwotsin/mimicry}} for different normalization methods on CIFAR-10 and CIFAR-100 datasets in Table \ref{table:normalization}. Among them, LN and GN obtained better performance than the most popular normalization method: Spectral normalization (mentioned in Section 4.2) and other methods significantly affect the stability of GANs training.
	
	\begin{table*}
		\caption{The summary of the layer normalization}
		\label{table:layer normalization}
		\scriptsize
		\centering
		\begin{tabular}{c|c | c | c  }	
			\toprule
			\midrule
			Method&{ Reference}& Classification&Inputs of $\gamma(c)$ and $\beta(c)$	 \\
			\hline
			Batch Normalization (BN)&2018 \cite{miyato2018spectral,xiang2017effects}& unconditional-based&-\\
			\hline
			Layer Normalization (LN)&2018 \cite{miyato2018spectral}& unconditional-based&-\\
			\hline
			Instance Normalization (IN)&2018 \cite{miyato2018spectral}& unconditional-based&-\\
			\hline
			Group Normalization (GN)&2018 \cite{wu2018group}& unconditional-based&-\\
			\hline
			Weight Normalization (WN) &2018 \cite{miyato2018spectral,xiang2017effects}& unconditional-based&-\\
			\hline
			Conditional Batch Normalization (CBN)&2018 \cite{miyato2018cgans,zhang2018self}&conditional-based&class label\\
			\hline
			Adaptive Instance Normalization (AdaIN)&2017 \cite{huang2017arbitrary},2019 \cite{karras2019style}&conditional-based&target images\\
			\hline
			Spatially-adaptive (de) Normalization (SPADE)&2019 \cite{park2019semantic}&conditional-based&sematic segmentation map\\
			\hline
			Attentive Normalization(AN)&2020 \cite{wang2020attentive}&conditional-based&self\\
			\bottomrule
		\end{tabular}
	\end{table*}
	
	\subsubsection{Conditional-based layer Normalization}
	Conditional-based layer normalization is only used for the generator of the conditional generation. It aims to introduce conditional information to each layer of the generator, which helps to improve the quality of the generated images. $\gamma(c)$ and $\beta(c)$ in Eq (44) are calculated with different features or class labels as input to the neural network in different methods. Miyato et al. \cite{miyato2018cgans} and Zhang et al. \cite{zhang2018self} used the Conditional Batch Normalization (CBN) to encode class labels, thereby improving the quality of conditional generation. Huang et al. \cite{huang2017arbitrary} and Karras et al. \cite{karras2019style} used the Adaptive Instance Normalization (AdaIN) with target images to improve the accuracy of style transfer. Park et al. \cite{park2019semantic} used the Spatially-Adaptive (de) Normalization (SPADE) with semantic segmentation image to incorporate semantic information into all layers. Wang et al. \cite{wang2020attentive} used the Attentive Normalization (AN) to model long-range dependent attention, which is similar to self-attention GAN \cite{zhang2018self}.
	
	In summary, the main difference between these conditional-based normalizations is the content of conditional inputs (c in Eq (45)). As the information of inputs is gradually enriched, the performance of conditional generation is gradually improved. The related summaries are shown in Table \ref{table:layer normalization}.
	
	\subsection{Inverse Gradient Penalty}
	Mode collapse is a common phenomenon in GANs' training, that is, changes in the latent space do not cause changes in the generated images. Geometrically, the phenomenon means that all the tangent vectors of the manifold are no longer independent of each other - some tangent vectors either disappear or become linearly correlated with each other. Intuitively, we can solve this problem by maximizing the Lipschitz constant of the generator, which is opposite of the gradient penalty of the discriminator described in the previous section. Based on this, inverse gradient penalty of the generator has been proposed. Concretely, under the little perturbation of the latent space, the generator needs to produce different images. Yang et al. \cite{yang2019diversity} use it in conditional generation, especially for tasks that are rich in conditional information, such as inpainting and super-resolution. 
	\begin{equation}
		\mathop{\max}\limits_{G}\mathcal{L}_z(G)=\mathop{\max}\mathbb{E}_{z_1,z_2}\left[\mathop{\min}\left(\frac{||G(y,z_1)-G(y,z_2)||}{||z_1-z_2||},\tau\right)\right],
	\end{equation}
	where $y$ is the class label and $\tau$ is the bound to ensure numerical stability. Unlike the intuition-based study described above, Odena et al. \cite{odena2018generator} demonstrate that the decreasing of singular value in the Jacobian matrix of the generator is the main reason for the mode collapse during GANs training. Furthermore, the singular value can be approximated by the gradient, so Jacobian clamping is used to limit singular values to $[\lambda_{min},\lambda_{\max}]$. The loss is expressed as:
	\begin{equation}
		\mathop{\min}\limits_{G}\mathcal{L}_z(G)=\big(\mathop{\max}(Q,\lambda_{\max})-\lambda_{\max}\big)^2
		+\big(\mathop{\min}(Q,\lambda_{\min})-\lambda_{\min}\big)^2,
	\end{equation}
	where $Q=||G(z)-G(z')||/||z-z'||$. 
	
	{In summary, the above two methods \cite{yang2019diversity,odena2018generator} are similar and mitigate the model collapse of generator to some extent. The key point is to improve the sensitivity of the generator to latent space. In addition to the above methods used to implement inverse gradient penalty for generators, some studies \cite{brock2016neural,brock2018large} adopt orthogonal regularization to enforce amenability to truncation by conditioning G to be
		smooth. Accordingly, the full space of $z$ will map to good output samples. Introducing orthogonality condition \cite{brock2016neural} is a direct method:}
	\begin{equation}
		R(W)=\beta\left\|W^{\top} W -I\right\|_{\mathrm{F}}^{2},
	\end{equation}
	{where W is a weight matrix and $\beta$ is a hyperparameter. However, this regularization is too limiting \cite{miyato2018spectral}. Therefore, a relaxed constraint has been designed by \cite{brock2018large}. Brock et al. \cite{brock2018large} apply Off-Diagonal Orthogonal Regularization (Off-Diagonal OR) to the generator directly enforcing the orthogonality condition:}
	\begin{equation}
		R_o(W)=\beta\left\|W^{\top} W \odot(\mathbf{1}-I)\right\|_{\mathrm{F}}^{2},
	\end{equation}
	{where $\mathbf{1}$ denotes a matrix with all elements set to 1. The Off-Diagonal OR makes G smooth so that the entire space of $z$ will map to good output samples. Orthogonality regularization is different from spectral normalization \cite{miyato2018spectral}. Orthogonality regularization destroys the information about the spectrum by setting all the singular values to one, while spectral normalization only makes the maximum singular be one.
	}

	\begin{table}
		\caption{ FID results for different normalization methods on CIFAR-10 and CIFAR-100 datasets. (  The structure is the same as SNGAN except that the discriminator uses different normalization methods).\tiny}
		\label{table:normalization}
		\centering
		\begin{tabular}{ccc}
			\toprule
			\midrule
			methods & CIFAR-10 & CIFAR-100 \\
			\hline
			None & 40.91 & 45.44 \\
			BN & 37.63 & 44.45\\
			LN & \textbf{19.21} & 21.15 \\
			IN & 34.14 & 43.64\\
			GN & 19.31 & $\textbf{20.80}$ \\
			WN & 24.28 & 29.96 \\
			SN&19.75&22.89\\
			\hline
		\end{tabular}
	\end{table}
	
	\section{Applications of Regularization and Normalization in SOTA GANs}
	{In this section, to provide a side view to the selection of regularization and normalization techniques, we investigate the applications of regularization and normalization techniques frequently employed in state-of-the-art and popular GANs. We select six methods (One per year from 2017-2022) categorized into two classes according to different tasks: Unconditional Generation and Conditional Generation. The selected methods and analysis are shown in Table \ref{table:SOTA }. PGGAN \cite{karras2017progressive} is a popular GAN model in recent years, which grows the size of both the generator and discriminator progressively. PGGAN empowers high-resolution image generation. Since PGGAN was proposed in 2017, only some simple regularization techniques were applied: WGAN-GP \cite{gulrajani2017improved}, BN \cite{miyato2018spectral}, and LN \cite{miyato2018spectral}; BigGAN \cite{brock2018large} is a popular conditional generative adversarial networks, which uses many regularization and normalization techniques, such as zc-GP\cite{mescheder2018training}, SN \cite{miyato2018spectral}, Off-Diagonal OR \cite{brock2018large}, and CBN \cite{miyato2018cgans}; AutoGAN \cite{gong2019autogan} is the first study introducing the Neural architecture search (NAS) to GANs. It defines the search space for the generator architecture and adopts Inception score as the reward to discover the best architecture. The main focus of AutoGAN is architecture, so AutoGAN only comprises SN \cite{miyato2018spectral}; StyleGAN2 \cite{karras2019style} is the most popular architecture of GANs, which produces photorealistic images with large varieties and is widely used in image generation, such as Image Completion \cite{zhao2021large}, Image-to-Image Translation \cite{richardson2020encoding}. StyleGAN2-ADA \cite{karras2020training} proposes a novelty adaptive data augmentation methods. Combining StyleGAN2 and adaptive data augmentation, StyleGAN2-ADA \cite{karras2020training} obtains impressive performance in image generation, particularly in data-effficient generation. Furthermore, InsGen \cite{yang2021data} combines StyleGAN2-ADA with contrastive learning, acquiring state of the art on many generation tasks and datasets. Recently, StyleGAN-XL \cite{sauer2022stylegan} scales StyleGAN to large diverse datasets and sets a new state-of-the-art on large-scale image synthesis. In summary, many regularization and normalization techniques have been used in state-of-the-art GANs with zc-GP and SN being more attractive to researchers. Data augmentation is a striking method and orthogonal to other ongoing researches on training, architecture, and regularization. Therefore, popular augmentation strategies, such as ADA, have been employed as default operations GANs training. Furthermore, self-supervision has been used to further improve the performance of GANs, which is also orthogonal to other methods.}
	
	\begin{table}
		\caption{The applications of the Regularization and Normalization techniques used in SOTA GANs.}
		\label{table:SOTA }
		\centering
		\footnotesize
		\begin{tabular}{c | c | c  |c | c | c|c }	
			\hline
			\hline
			{ Method}& Task&\makecell[c]{Gradient \\Penalty}&\makecell[c]{Data augmentation\\ and preprocessing}&\makecell{Self\\supervision}&\makecell[c]{Weight\\ normalization }&\makecell[c]{Layer\\ normalization}\\
			\hline
			\makecell{PGGAN\\(2017\cite{karras2017progressive})}&\makecell{Unconditinal\\ Generation}&WGAN-GP&\makecell{None}&None&\makecell{None}&BN: G,LN: D\\
			\hline
			\makecell{BigGAN\\ (2018 \cite{brock2018large})}&\makecell{Conditinal \\Generation}&zc-GP&None&None&\makecell{SN: G, D}&CBN\\
			\hline
			\makecell{AutoGAN\\ (2019 \cite{gong2019autogan})}&\makecell{Unconditinal\\ Generation}&None&None&None&\makecell{SN:D}&None\\
			\hline
			\makecell{StyleGAN2-ADA\\(2020\cite{karras2020training})}&\makecell{Unconditinal \\Generation}&zc-GP&\makecell{Adaptive}&None&None&IN\\
			\hline
			\makecell{InsGen\\(2021\cite{yang2021data})}&\makecell{Unconditinal \\Generation}&zc-GP&\makecell{Adaptive}&contrastive&None&IN\\
			\hline
			\makecell{StyleGAN-XL\\(2022\cite{sauer2022stylegan})}&\makecell{Conditinal \\Generation}&None&\makecell{Translation\\Cutout}&None&SN:D&IN\\
			\hline
		\end{tabular}
	\end{table}

	\section{Summary and Outlook}
	\subsection{Summary}
	Recently, significant achievements of GANs have been made in generation tasks and the network has been widely used in many computer vision tasks, such as image inpainting, style transfer, text-to-image translations, and attribute editing. However, due to the overconfident assumptions, the training faces many challenges, such as non-convergence, mode collapse, gradient vanishing, and overfitting. To mitigate these problems, many solutions focus on designing new architectures, new loss functions, new optimization methods, and regularization and normalization techniques. 
	
	In this paper, we study GANs training from three perspectives and propose a new taxonomy, denoted as \textbf{"Training dynamics"}, \textbf{"Fitting distribution"}, \textbf{"Real \& Fake"}, and \textbf{"Other methods"}, to survey the different regularization and normalization techniques during GANs training. Our study provides a systematic and comprehensive analysis of the reviewed methods to serve researchers of the community. In addition, we also demonstrate the motivation and objectives of different methods and compare the performance of some popular methods in a fair manner quantitatively, which has implications for future research in selecting their research topics or developing their approaches.
	\subsection{Outlook}
	By reviewing the regularization and normalization of GANs, the following questions and thoughts are proposed based on different perspectives of GANs training:
	\begin{itemize}
		\item [1)] 
		\textit{What is a good distance metric, and which divergence should be used in GANs training?} The priority in the training process of GANs is to find a suitable divergence to measure the distance between the generated distribution and the true distribution. Wasserstein divergence is important for the training of GANs. However, it is uncertain whether the next proposed divergence performs better.
		\item [2)]
		\textit{What is the main difference between real images and generated images?} During the training of unconstrained and unprioritized GANs, if we can quantitatively represent the difference between real images and generated images from different perspectives, the efficient regularization methods can be designed based on this.
		\item [3)]
		\textit{How to avoid real images forgetting\footnote{Real images forgetting is caused by not introducing real images while training the generator, which is different from discriminator forgetting.}?} As acknowledged, real images do not directly participate in the training of the generator, thus the discriminator needs to remember the characteristics of the real images to optimize the generator indirectly. We call this the real images forgetting. We conjecture that real images forgetting may exist, and which may increase the difficulty of GANs training. Some works might serve as basis to prove this hypothesis and propose effective solutions.
		\item [4)]
		Recent studies show that discriminator suffers from overfitting and discriminator forgetting. It is a common problem of neural networks, which is caused by the shortcut of the loss driven method. Some new methods, such as contrastive learning, representation learining, can be proposed to improve the generalization of the discriminator.
		\item [5)]
		Recently, diffusion model \cite{dhariwal2021diffusion} acquires the impressive performance in image generation. One possible reason for success is the phased training strategy in diffusion model \cite{dhariwal2021diffusion}. Inspired by this, some strategies to reduce the difficulty of GANs training may be proposed.
	\end{itemize}

	\section*{Acknowledgment}
	
	The work is partially supported by the National Natural Science Foundation of China under Grand No.U19B2044, No.61836011 and No.91746209. We are very grateful to the help of Jianlin Su, whose blog is \url{https://spaces.ac.cn/tag/GAN/}.
	
	\bibliographystyle{ACM-Reference-Format}
	\bibliography{bare_jrnl_transmag}
	
	\appendix
	\clearpage
	\section{Supplementary Online-only Material}
	\subsection{Optimal Transport and Lipschitz Continuity}
	\label{sect:A-1}
	Optimal transport \cite{bonnotte2013knothe} was proposed in the 18th century to minimize the transportation cost while preserving the measure quantities. Given the space with probability measures $(X,\mu)$ and $(Y,\upsilon)$, if there is a map $T:X\rightarrow Y$ which is measure-preserving, then for any $B\subset Y$, having:
	\begin{equation}\label{eqn1}
		\int_{T^{-1}(B)}\mathrm{d}\mu(x)=\int_B \mathrm{d}\upsilon(y).
	\end{equation}
	Writing the measure-preserving map as $T_*(\mu)=\upsilon$. For any $x\in X$ and $y\in Y$, the transportation distance is defined as $c(x,y)$, the total transportation cost is given by:
	\begin{equation}\label{eqn1}
		C(T):=\int_X c(x,T(x)) \mathrm{d}\mu(x).
	\end{equation}
	In the 18th century, Monge et al. \cite{monge1781memoire} proposed the Optimal Mass Transportation Map that corresponds to the smallest total transportation cost: $C(T)$. The transportation cost corresponding to the optimal transportation map is called the Wasserstein distance between probability measures $\mu$ and $\upsilon$:
	\begin{equation}\label{eqn3}
		W_c(\mu,\upsilon)=\mathop{\min}\limits_{T}\left\{\int_X c(x,T(x)) \mathrm{d}\mu(x)\ |\ T_*(\mu)=\upsilon\right\}.
	\end{equation}
	In 1940s, Kantorovich \cite{kantorovich2006problem} proved the existence and uniqueness of the solution for Monge problem, and according to the duality of linear programming, the Kantorovich-Rubinstein (KR) duality of Wasserstein distance is given by:
	\begin{equation}\label{eqn1}
		W_c(\mu,\upsilon)
		=\mathop{\max}\limits_{\varphi ,\psi}\left\{\int_X \varphi \mathrm{d}\mu\ +\int_Y \psi \mathrm{d}\upsilon\ |\ \varphi(x)+\psi(y)\leq c(x,y)\right\}.
	\end{equation}
	This dual problem is constrained, defining the c-transform: $\psi(y)=\varphi^c(y):=inf_x\{c(x,y)-\varphi(x)\}$, and the Wasserstein distance becomes:
	\begin{equation}\label{eqn1}
		W_c(\mu,\upsilon)=\mathop{\max}\limits_{\varphi}\left\{\int_X \varphi \mathrm{d}\mu\ +\int_Y \varphi^c \mathrm{d}\upsilon\right\},
	\end{equation}
	where $\varphi$ is called the Kantorovich potential. It can be shown that if $c(x,y)=|x-y|$ and Kantorovich potential satisfies the 1-Lipschitz continuity, then $\varphi^c=-\varphi$. Kantorovich potential can be fitted by a deep neural network, which is recorded as $\varphi_\xi$. Wasserstein distance is:
	\begin{equation}
		W_c(\mu,\upsilon)=\mathop{\max}\limits_{||\varphi_\xi||_L\leq 1}\left\{\int_X \varphi_\xi \mathrm{d}\mu\ -\int_Y \varphi_\xi \mathrm{d}\upsilon\right\}.
		\label{eqn7}
	\end{equation}
	If $X$ is the generated image space, $Y$ is the real sample space, $Z$ is latent space and $g_\theta$ is the geneartor, the Wasserstein GANs (WGAN) is formulated as a min-max problem:
	\begin{equation}\label{eqn1}
		\mathop{\min}\limits_{\theta}\mathop{\max}\limits_{||\varphi_\xi||_L\leq 1}\left\{\int_Z \varphi_\xi(g_\theta(z)) \mathrm{d}z \ -\int_Y \varphi_\xi(y) \mathrm{d}y\right\}.
	\end{equation}
	In the optimization process, the generator and the Kantorovich potential function (discriminator) are independent of each other, optimized in a step-by-step iteration.
	
	If $c(x,y)=\frac{|x-y|^2}{2}$, there is a convex function $u$ that is called Brenier potential \cite{brenier1991polar}. The optimal transportation map is given by the gradient map of Brenier potential: $T(x)=\nabla u(x)$. There exists a relationship between Kantorovich potential and Brenier potential \cite{lei2019geometric}: 
	\begin{equation}\label{eqn1}
		u(x)=\frac{|x|^2}{2}-\varphi(x).
	\end{equation}
	From the previous discussion, it is evident that the optimal transportation map (Brenier potential) corresponds to the generator, and Kantorovich potential corresponds to the discriminator. After the discriminator is optimized, the generator is directly drivable without the optimization process \cite{lei2019geometric}.
	
	The transportation cost of Eq (\ref{eqn3}) is defined as the form of two distribution distances:
	\begin{equation}\label{eqn10}
		OT(P||Q)=\mathop{inf}\limits_{\pi}\int\pi(x,y)c(x,y)\mathrm{d}x\mathrm{d}y,
	\end{equation}
	where $\pi(x,y)$ is the joint distribution, satisfying $\int_y\pi(x,y)dy=P(x)$ and $\int_x\pi(x,y)dx=Q(y)$. The dual form of Eq (\ref{eqn10}) is derived as follows::
	\begin{equation}\label{eqn1}
		OT(P||Q)=\mathop{\max}\limits_{\varphi ,\psi}\{\int_x \varphi(x)P(x) \mathrm{d}x\ \\ +\int_y\psi(y)Q(y) \mathrm{d}y\ |\ \varphi(x)+\psi(y)\leq c(x,y)\}.
	\end{equation} 
	Considering the optimal transportation with regular terms, Peyr{\'e} et al. \cite{peyre2019computational} added the entropic regularization for optimal transportation that transforms the dual problem into a smooth unconstrained convex problem. The regularized optimal transport is defined as:
	\begin{equation}\label{eqn12}
		OT_c(P||Q)=\mathop{\min}\limits_{\pi}\int\pi(x,y)c(x,y)\mathrm{d}x\mathrm{d}y+\epsilon E(\pi).
	\end{equation}
	If  $E(\pi)=\int_x\int_y\pi(x,y)\log(\frac{\pi(x,y)}{P(x)Q(y)})\mathrm{d}x\mathrm{d}y$, Eq (\ref{eqn12}) can be written as:
	\begin{equation}\label{eqn13}
		\begin{split}
			OT_c(P||Q)=&\mathop{\min}\limits_{\pi}\int\pi(x,y)c(x,y)\mathrm{d}x\mathrm{d}y+\epsilon \int_x\int_y\pi(x,y)\log\left(\frac{\pi(x,y)}{P(x)Q(y)}\right)\mathrm{d}x\mathrm{d}y\\
			s.t. \int_y\pi(x,y)&\mathrm{d}y=P(x),\int_x\pi(x,y)\mathrm{d}x=Q(y).
		\end{split}
	\end{equation}
	The dual form of Eq (\ref{eqn13}) becomes:
	\begin{equation}
		\begin{split}
			OT_c(P||Q)
			&=\mathop{\max}\limits_{\varphi ,\psi}\int_x \varphi(x)P(x) \mathrm{d}x\ +\int_y\psi(y)Q(y)\mathrm{d}y\\
			&+\frac{\epsilon}{e}\int_x\int_y\exp\left(\frac{-\left(c(x,y)+\varphi(x)+\psi(y)\right)}{\epsilon}\right)\mathrm{d}x\mathrm{d}y.
		\end{split}
		\label{EQ:eqn14}
	\end{equation}
	Petzka et al. \cite{petzka2017regularization} set $c(x,y)=||x-y||_2$ in Eq (\ref{EQ:eqn14}), and the dual form of optimal transport with the regular term can be expressed as:
	\begin{equation}
		\mathop{\sup}\limits_{\varphi,\psi}\{\mathbb{E}_{x\sim{p(x)}}[\varphi(x)]-\mathbb{E}_{y\sim q(y)}[\psi(y)]
		-\frac{4}{\epsilon}\int\int\mathop{\max}\{0,(\varphi(x)-\psi(y)-||x-y||_2)\}^2\mathrm{d}p(x)\mathrm{d}q(y) \}.
		\label{EQ:eqn32}
	\end{equation}
	Similar to dealing with a single function, one can replace $\varphi = \psi$ in Eq (\ref{EQ:eqn32}), which leads to the objective of minimum:
	\begin{equation}\label{eqn1}
		\mathbb{E}_{y\sim q(y)}[\varphi(y)]-\mathbb{E}_{x\sim p(x)}[\varphi(x)]
		+\frac{4}{\epsilon}\int\int\mathop{\max}\{0,(\varphi(x)-\varphi(y)-||x-y||_2)\}^2\mathrm{d}p(x)\mathrm{d}q(y).
	\end{equation}

	\subsection{Some Propositions of Training Dynamic in GANs}
	\label{sect:A-2}
	\newenvironment{lemma1}{{\indent\it \textbf{Proposition 2.1:}}}{\hfill \par}
	\begin{lemma1}
		\textit{For zero-sum games, $v^{'}$is negative semi-definite for any local Nash-equilibrium. Conversely, if $v(\bar{x})=0$ and $v^{'}$is negative definite, then $\bar{x}$ is a local Nash-equilibrium.
		}
	\end{lemma1}
	\newenvironment{proof1}{{\indent\it Proof 2.1:}}{\hfill $\square$\par}
	\begin{proof1}
		Refer to \cite{mescheder2017numerics}
	\end{proof1}
	\textit{Proposition 2.1} \cite{mescheder2017numerics} gives the conditions for the local convergence of GANs, which is converted into the negative semi-definite problem of the Jacobian matrix. Negative semi-definite of the Jacobian matrix corresponds to its eigenvalue less than or equal to 0. If the eigenvalue of the Jacobian matrix at a certain point is a negative real number, the training process can converge; but if the eigenvalue is complex and the real part of the eigenvalue is small and the imaginary part is relatively large, the training process is difficult to converge unless the learning rate is very small.

	\newenvironment{lemma2}{{\indent\it \textbf{Proposition 2.2:}}}{\hfill \par}
	\begin{lemma2}
		\textit{
			Let $F:\Omega\rightarrow\Omega$ be a continuously differentiable function on an open subset $\Omega$ of $R^n$ and let $\bar{x}\in\Omega$ be so that: 1. $F(\bar{x})=\bar{x}$ and 2. the absolute values of the eigenvalues of the Jacobian $F^{'}(x)$ are all smaller than 1.}
		
		\textit{There is an open neighborhood $U$ of $\bar{x}$ so that for all $x_0\in U$, the iterates $F^{(k)}(x_0)$ converge to $\bar{x}$. The rate of convergence is at least linear. More precisely, the error $||F^{(k)}(x_0)-\bar{x}||$ is in $\mathcal{O}(|\lambda_{max}|^k)$ for $k\rightarrow\infty$ where $\lambda_{max}$ is the eigenvalue of $F^{'}(\bar{x})$ with the largest absolute value.}
	\end{lemma2}
	\newenvironment{proof2}{{\indent\it Proof 2.2:}}{\hfill $\square$\par}
	\begin{proof2}
		Refer to Section 3 in \cite{mescheder2017numerics} and Proposition 4.4.1 in \cite{mangasarian1994nonlinear}.
	\end{proof2}
	
	\subsection{Spectral Norm and the Lipschitz Constant}
	\label{sect:A-3}
	
	1-Lipschitz continuity is represented as:
	\begin{equation}
		||D(x_1)-D(x_2)||\leq ||x_1-x_2||.\footnote{Lipschitz continuity can be defined by any form of norm.}
	\end{equation}
	Generally, considering the K-Lipschitz for a neural network $f(x)$:
	\begin{equation}
		f(x)=g_N\circ\cdots g_2\circ g_1(x),\footnote{$\circ$ is the symbol for function cascade. Specifically, $h\circ g(x)=h(g(x))$. This definition of neural network is not general, such as DenseNet \cite{Huang_2017_CVPR} and ResNet \cite{he2016deep}, which can not be defined like this. Therefore, we do not strictly derive the relationship between the matrix norm and Lipschitz continuity.}
	\end{equation}
	where $g_i(x)=\sigma (W_i x+b_i)$. And K-Lipschitz continuity for $f(x)$ is:
	\begin{equation}
		||f(x_1)-f(x_2)||\leq \mathrm{K}||x_1-x_2||,
		\label{EQ:eqn17}
	\end{equation}
	where K is Lipschitz constant of the function $f$. Due to the consistency of Lipschitz $||h\circ g||_{Lip}\leq ||h||_{Lip}\cdot||g||_{Lip}$, $g_i$ needs to satisfy the C-Lipschitz continuity ($\mathrm{C}=\sqrt[N]{\mathrm{K}}$) so that $f$ satisfies the K-Lipschitz continuity:
	\begin{equation}
		||g_i(x_1)-g_i(x_2)||\leq \mathrm{C}||x_1-x_2||,
	\end{equation}
	\begin{equation}
		||\sigma(Wx_1+b)-\sigma(Wx_2+b)||\leq \mathrm{C}||x_1-x_2||.
		\label{eq:23}
	\end{equation}
	When $x_1\rightarrow x_2$, the Taylor expansion of Eq (\ref{eq:23}):
	\begin{equation}
		||\frac{\partial\sigma}{\partial x} W(x_1-x_2)||\leq \mathrm{C}||x_1-x_2||.
	\end{equation}
	Normally, $\sigma$ is a function with limited derivatives such as Sigmoid, so the $\mathrm{C'}$-Lipschitz continuity is be written as:
	\begin{equation}
		|| W(x_1-x_2)||\leq \mathrm{C'}||x_1-x_2||,
	\end{equation}
	where $\mathrm{C'}$ is a limited constant, which is determined by $\frac{\partial\sigma}{\partial x}$ and $\mathrm{C}$.
	Similarly, the spectral norm of matrix is defined by:
	\begin{equation}
		||W||_2=\mathop{\max}\limits_{x\not=0}\frac{||Wx||}{||x||}.
	\end{equation}
	In this context, the spectral norm $||W||_2$ can be used to represent the Lipschitz constant $\mathrm{C'}$. The Lipschitz continuity is achieved by normalizing the spectral norm of the weight, approximately.	
	

\end{document}